%% file: memento-S-fancy.tex
\documentclass[11pt,a4paper]{article}

\usepackage[utf8]{inputenc}
\usepackage[T1]{fontenc}
\usepackage{lmodern}
\usepackage[margin=2.5cm]{geometry}
\usepackage{amsmath,amssymb,amsthm}
\usepackage{bm}
\usepackage{algorithm}
\usepackage{algorithmic}
\usepackage{graphicx}
\graphicspath{{figure/}}
\usepackage{xcolor}
\usepackage[most]{tcolorbox}
\usepackage{tikz}
\usetikzlibrary{shapes.callouts,positioning,shadows,calc,arrows.meta,decorations.pathreplacing,
  patterns,fit,backgrounds,matrix,chains,scopes}
\usepackage{enumitem}
\usepackage{hyperref}
\usepackage[numbers]{natbib}
\usepackage{fontawesome5}
\usepackage{mdframed}
\usepackage{fancyhdr}
\usepackage{etoolbox}
\usepackage{titlesec}
\usepackage{booktabs}
\usepackage{multicol}
\usepackage{caption}
\usepackage{microtype}
\usepackage{listings}
\usepackage{upquote}
\usepackage{accsupp}
\usepackage{pgfplots}
\pgfplotsset{compat=1.18}
\usepackage{subcaption}
\usepackage{colortbl}
\usepackage{multirow}
\usepackage{pifont}
\usepgfplotslibrary{groupplots}
\usepackage{svg}
\definecolor{lvl1}{RGB}{31,119,180}
\definecolor{lvl2}{RGB}{255,127,14}
\definecolor{lvl3}{RGB}{44,160,44}
\definecolor{lvlA}{RGB}{214,39,40}

\usepackage[T1]{fontenc}
\usepackage{lmodern}
\usepackage{amsmath,amssymb}
\input{glyphtounicode}
\input{macro}

\definecolor{titleRed}{HTML}{B71C1C}
\definecolor{theoryBG}{HTML}{FFF8E7}
\definecolor{theoryBorder}{HTML}{D4A843}
\definecolor{practiceBG}{HTML}{E8F4FD}
\definecolor{practiceBorder}{HTML}{3B82B6}
\definecolor{bridgeBG}{HTML}{F0FFF0}
\definecolor{bridgeBorder}{HTML}{2E8B57}
\definecolor{dialogueBG}{HTML}{FAF0FA}
\definecolor{dialogueBorder}{HTML}{8B5E8B}
\definecolor{sharedBG}{HTML}{FAFAFA}
\definecolor{marginJ}{HTML}{B71C1C}
\definecolor{marginH}{HTML}{E65100}
\definecolor{marginS}{HTML}{1565C0}
\definecolor{trackR}{HTML}{B8860B}
\definecolor{trackP}{HTML}{2166AC}
\definecolor{epilogueBG}{HTML}{FFF5F5}
\definecolor{codeBG}{HTML}{1E1E2E}
\definecolor{codeFrame}{HTML}{45475A}
\definecolor{codeGreen}{HTML}{A6E3A1}
\definecolor{codeBlue}{HTML}{89B4FA}
\definecolor{codePurple}{HTML}{CBA6F7}
\definecolor{codeYellow}{HTML}{F9E2AF}
\definecolor{codeRed}{HTML}{F38BA8}
\definecolor{codeOrange}{HTML}{FAB387}
\definecolor{codeComment}{HTML}{6C7086}
\definecolor{warningBG}{HTML}{FFF3E0}
\definecolor{warningBorder}{HTML}{E65100}
\definecolor{insightBG}{HTML}{E8F5E9}
\definecolor{insightBorder}{HTML}{2E7D32}
\definecolor{keyresultBG}{HTML}{FCE4EC}
\definecolor{keyresultBorder}{HTML}{C62828}
\definecolor{pseudoBG}{HTML}{F5F5FF}
\definecolor{pseudoBorder}{HTML}{5C6BC0}

\lstdefinestyle{darkpython}{
  language=Python,
  backgroundcolor=\color{codeBG},
  basicstyle=\ttfamily\small\color{white},
  keywordstyle=\color{codePurple}\bfseries,
  stringstyle=\color{codeGreen},
  commentstyle=\color{codeComment}\itshape,
  numberstyle=\tiny\color{codeComment},
  identifierstyle=\color{codeBlue},
  emph={Agent,Memory,Environment,RetrievalPolicy,ParzenKernel,SoftPolicyIterator,EpisodicMemory,
    ReflectedMDP,ReadWriteLoop,VoidCase,MemoryConsolidator},
  emphstyle=\color{codeYellow},
  emph={[2]self,True,False,None},
  emphstyle={[2]\color{codeOrange}},
  numbers=left,
  numbersep=8pt,
  frame=single,
  rulecolor=\color{codeFrame},
  framesep=6pt,
  xleftmargin=18pt,
  framexleftmargin=18pt,
  showstringspaces=false,
  tabsize=4,
  breaklines=true,
  breakatwhitespace=true,
  aboveskip=8pt,
  belowskip=8pt,
  literate={->}{{{\color{codeRed}->}}}2
           {>=}{{{\color{codeRed}>=}}}2
           {<=}{{{\color{codeRed}<=}}}2
           {!=}{{{\color{codeRed}!=}}}2
           {==}{{{\color{codeRed}==}}}2
           {**}{{{\color{codeRed}**}}}2,
}

\lstdefinestyle{darkyaml}{
  basicstyle=\ttfamily\small\color{white},
  backgroundcolor=\color{codeBG},
  keywordstyle=\color{codeYellow},
  stringstyle=\color{codeGreen},
  commentstyle=\color{codeComment}\itshape,
  frame=single,
  rulecolor=\color{codeFrame},
  framesep=6pt,
  xleftmargin=12pt,
  framexleftmargin=12pt,
  showstringspaces=false,
  breaklines=true,
  aboveskip=8pt,
  belowskip=8pt,
  morecomment=[l]{\#},
}

\lstdefinestyle{darkbash}{
  language=bash,
  basicstyle=\ttfamily\small\color{white},
  backgroundcolor=\color{codeBG},
  keywordstyle=\color{codeGreen}\bfseries,
  commentstyle=\color{codeComment}\itshape,
  stringstyle=\color{codeYellow},
  frame=single,
  rulecolor=\color{codeFrame},
  framesep=6pt,
  xleftmargin=12pt,
  framexleftmargin=12pt,
  showstringspaces=false,
  breaklines=true,
  aboveskip=8pt,
  belowskip=8pt,
  literate={~}{{\raise.35ex\hbox{$\scriptstyle\sim$}}}1
           {\$}{{\BeginAccSupp{method=escape,ActualText=}\color{codeGreen}\$\EndAccSupp{}}}1
           {-}{{-}}1,
}

\newtcolorbox{theorytrack}[1][]{%
  enhanced, colback=theoryBG, colframe=theoryBorder, boxrule=1.2pt,
  left=8pt,right=8pt,top=8pt,bottom=8pt,
  fonttitle=\bfseries\sffamily\color{trackR},
  title={\faFlask~\textsf{RESEARCH TRACK}~~#1},
  attach boxed title to top left={xshift=6pt,yshift=-3mm},
  boxed title style={colback=theoryBG,colframe=theoryBorder,boxrule=0.8pt},
  arc=2pt, shadow={1pt}{-1pt}{0pt}{black!15}, breakable
}

\newtcolorbox{practicetrack}[1][]{%
  enhanced, colback=practiceBG, colframe=practiceBorder, boxrule=1.2pt,
  left=8pt,right=8pt,top=8pt,bottom=8pt,
  fonttitle=\bfseries\sffamily\color{trackP},
  title={\faCode~\textsf{PRACTITIONER TRACK}~~#1},
  attach boxed title to top left={xshift=6pt,yshift=-3mm},
  boxed title style={colback=practiceBG,colframe=practiceBorder,boxrule=0.8pt},
  arc=2pt, shadow={1pt}{-1pt}{0pt}{black!15}, breakable
}

\newtcolorbox{bridgebox}[1][]{%
  enhanced, colback=bridgeBG, colframe=bridgeBorder, boxrule=1.5pt,
  left=8pt,right=8pt,top=8pt,bottom=8pt,
  fonttitle=\bfseries\sffamily\color{bridgeBorder},
  title={\faLink~\textsf{BRIDGE}~~#1},
  attach boxed title to top center={yshift=-3mm},
  boxed title style={colback=bridgeBG,colframe=bridgeBorder,boxrule=0.8pt},
  arc=3pt, shadow={1.5pt}{-1.5pt}{0pt}{black!12}, breakable
}

\newtcolorbox{dialogue}[1][]{%
  enhanced, colback=dialogueBG, colframe=dialogueBorder, boxrule=1.5pt,
  left=10pt,right=10pt,top=10pt,bottom=10pt,
  fonttitle=\bfseries\sffamily\large\color{dialogueBorder},
  title={\faTheaterMasks~\textsf{THE LOBBY}~~#1},
  attach boxed title to top center={yshift=-3mm},
  boxed title style={colback=dialogueBG,colframe=dialogueBorder,boxrule=0.8pt},
  arc=4pt, shadow={2pt}{-2pt}{0pt}{black!15}, breakable
}

\newtcolorbox{epilogue}[1][]{%
  enhanced, colback=epilogueBG, colframe=dialogueBorder!60, boxrule=1pt,
  left=10pt,right=10pt,top=8pt,bottom=8pt,
  fonttitle=\bfseries\sffamily\color{dialogueBorder!80},
  title={\faStar~\textsf{EPILOGUE}~~#1},
  attach boxed title to top center={yshift=-3mm},
  boxed title style={colback=epilogueBG,colframe=dialogueBorder!60,boxrule=0.8pt},
  arc=3pt, breakable
}

\newtcolorbox{chaptermap}{%
  enhanced, colback=sharedBG, colframe=black!40, boxrule=1pt,
  left=8pt,right=8pt,top=8pt,bottom=8pt,
  fonttitle=\bfseries\sffamily,
  title={\faMap~\textsf{READING MAP}},
  attach boxed title to top left={xshift=6pt,yshift=-3mm},
  boxed title style={colback=sharedBG,colframe=black!40,boxrule=0.8pt},
  arc=2pt, breakable
}

\newtcolorbox{warningbox}[1][]{%
  enhanced, colback=warningBG, colframe=warningBorder, boxrule=1pt,
  left=8pt,right=8pt,top=6pt,bottom=6pt,
  fonttitle=\bfseries\sffamily\color{warningBorder},
  title={\faExclamationTriangle~\textsf{WARNING}~~#1},
  attach boxed title to top left={xshift=6pt,yshift=-3mm},
  boxed title style={colback=warningBG,colframe=warningBorder,boxrule=0.8pt},
  arc=2pt, breakable
}

\newtcolorbox{insightbox}[1][]{%
  enhanced, colback=insightBG, colframe=insightBorder, boxrule=1pt,
  left=8pt,right=8pt,top=6pt,bottom=6pt,
  fonttitle=\bfseries\sffamily\color{insightBorder},
  title={\faLightbulb~\textsf{KEY INSIGHT}~~#1},
  attach boxed title to top left={xshift=6pt,yshift=-3mm},
  boxed title style={colback=insightBG,colframe=insightBorder,boxrule=0.8pt},
  arc=2pt, breakable
}

\newtcolorbox{keyresult}[1][]{%
  enhanced, colback=keyresultBG, colframe=keyresultBorder, boxrule=1.2pt,
  left=8pt,right=8pt,top=6pt,bottom=6pt,
  fonttitle=\bfseries\sffamily\color{keyresultBorder},
  title={\faTrophy~\textsf{KEY RESULT}~~#1},
  attach boxed title to top left={xshift=6pt,yshift=-3mm},
  boxed title style={colback=keyresultBG,colframe=keyresultBorder,boxrule=0.8pt},
  arc=2pt, breakable
}

\newtcolorbox{pseudobox}[1][]{%
  enhanced, colback=pseudoBG, colframe=pseudoBorder, boxrule=1pt,
  left=8pt,right=8pt,top=6pt,bottom=6pt,
  fonttitle=\bfseries\sffamily\color{pseudoBorder},
  title={\faCogs~\textsf{ALGORITHM}~~#1},
  attach boxed title to top left={xshift=6pt,yshift=-3mm},
  boxed title style={colback=pseudoBG,colframe=pseudoBorder,boxrule=0.8pt},
  arc=2pt, breakable
}

\newtcolorbox{sharedtrack}[1][]{%
  enhanced,
  colback=sharedBG,
  colframe=black!40,
  boxrule=1.2pt,
  left=8pt,right=8pt,top=8pt,bottom=8pt,
  fonttitle=\bfseries\sffamily\color{black!70},
  title={\faLayerGroup~\textsf{SHARED TRACK}~~#1},
  attach boxed title to top left={xshift=6pt,yshift=-3mm},
  boxed title style={
    colback=sharedBG,
    colframe=black!40,
    boxrule=0.8pt
  },
  arc=2pt,
  shadow={1pt}{-1pt}{0pt}{black!12},
  breakable
}

\newtcolorbox{inlinenote}[2][]{%
  enhanced, colback=#2!8, colframe=#2!50, boxrule=0.6pt,
  left=6pt,right=6pt,top=3pt,bottom=3pt, arc=2pt,
  fontupper=\footnotesize\sffamily,
  before skip=4pt, after skip=4pt, #1
}
\newcommand{\profJ}[1]{%
  \begin{inlinenote}{marginJ}%
  \textbf{\color{marginJ}\faGraduationCap~J:}~\textit{#1}%
  \end{inlinenote}%
}
\newcommand{\studentH}[1]{%
  \begin{inlinenote}{marginH}%
  \textbf{\color{marginH}\faFire~H:}~\textit{#1}%
  \end{inlinenote}%
}
\newcommand{\seniorS}[1]{%
  \begin{inlinenote}{marginS}%
  \textbf{\color{marginS}\faServer~S:}~\textit{#1}%
  \end{inlinenote}%
}

\newcommand{\J}[0]{{\sffamily\bfseries\color{marginJ}J}}
\newcommand{\Hstudent}[0]{{\sffamily\bfseries\color{marginH}H}}
\newcommand{\Senior}[0]{{\sffamily\bfseries\color{marginS}S}}
\newcommand{\stage}[1]{{\small\itshape\color{black!50}(#1)}}

\newcommand{\Rtag}{{\sffamily\bfseries\colorbox{theoryBG}{\color{trackR}\,R\,}}}
\newcommand{\Ptag}{{\sffamily\bfseries\colorbox{practiceBG}{\color{trackP}\,P\,}}}
\newcommand{\Stag}{{\sffamily\bfseries\colorbox{sharedBG}{\color{black!60}\,S\,}}}

\fancypagestyle{empty}{%
  \fancyhf{}%
}
\fancypagestyle{plain}{%
  \fancyhf{}%
}

\begin{document}

\begin{titlepage}
\centering

{\LARGE\bfseries\sffamily\color{titleRed} Memento-Skills: Let Agents Design Agents}
\\[20pt]

{\large \textbf{Memento-Team}}


\begin{abstract}
\noindent
We introduce \emph{Memento-Skills}, a generalist, continually-learnable LLM agent system that functions as an \emph{agent-designing agent}: it autonomously constructs, adapts, and improves task-specific agents through experience. The system is built on a memory-based reinforcement learning framework with \emph{stateful prompts}, where reusable skills (stored as structured markdown files) serve as persistent, evolving memory. These skills encode both behaviour and context, enabling the agent to carry forward knowledge across interactions.

Starting from simple elementary skills (like Web search and terminal operations), the agent continually improves via the \emph{Read--Write Reflective Learning} mechanism introduced in \emph{Memento~2}~\cite{wang2025memento2}. In the \emph{read} phase, a behaviour-trainable skill router selects the most relevant skill conditioned on the current stateful prompt; in the \emph{write} phase, the agent updates and expands its skill library based on new experience. This closed-loop design enables \emph{continual learning without updating LLM parameters}, as all adaptation is realised through the evolution of externalised skills and prompts.

Unlike prior approaches that rely on human-designed agents, Memento-Skills enables a generalist agent to \emph{design agents end-to-end} for new tasks. Through iterative skill generation and refinement, the system progressively improves its own capabilities. Experiments on the \emph{General AI Assistants} benchmark and \emph{Humanity's Last Exam} demonstrate sustained gains, achieving 26.2\% and 116.2\% relative improvements in overall accuracy, respectively. Code is available at \url{https://github.com/Memento-Teams/Memento-Skills}.
\end{abstract}

\begin{figure}[h!]
    \centering
    \includegraphics[width=1.1\linewidth]{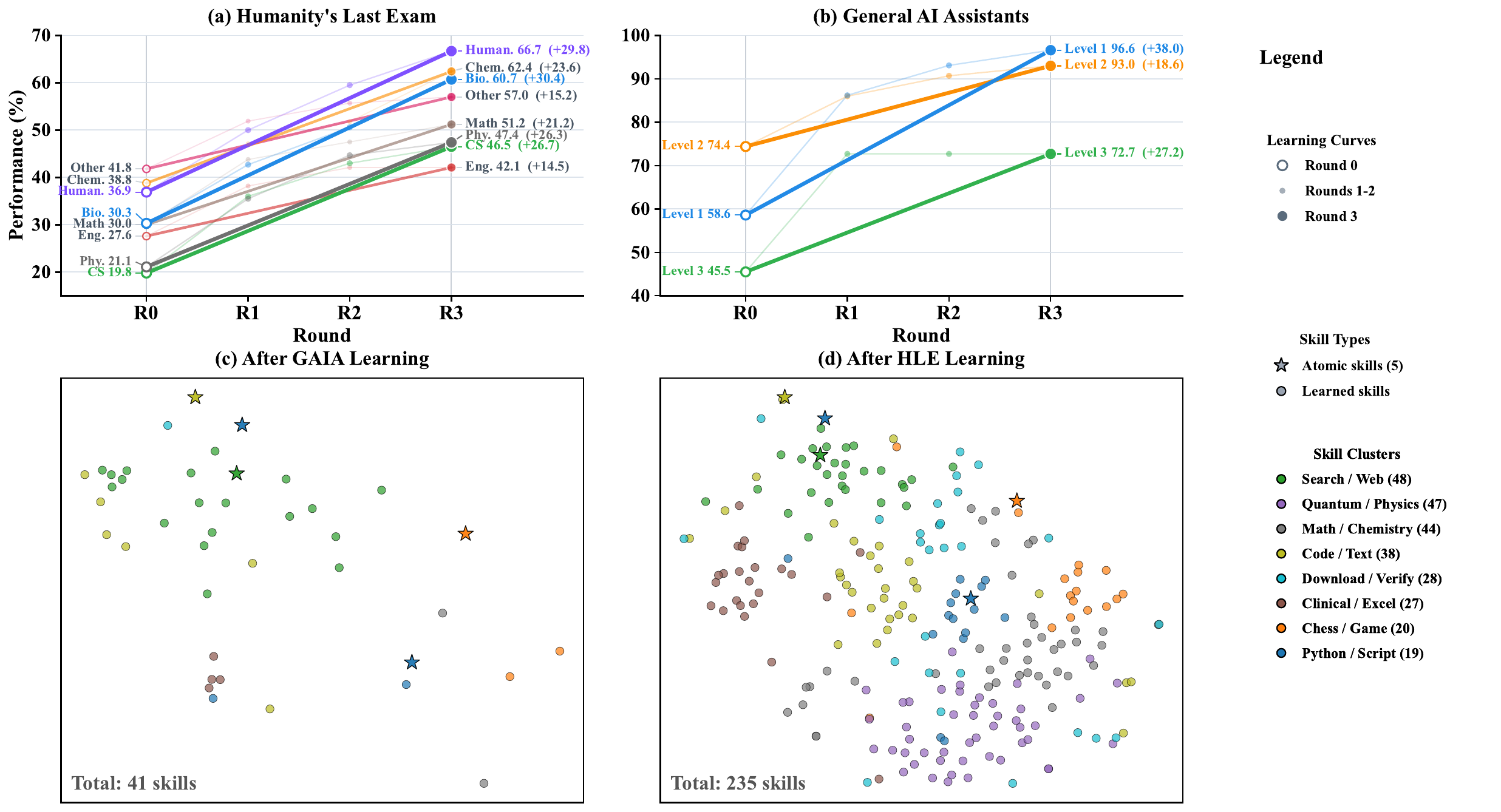}
    \caption{Overview of self-evolving results of Memento-Skills on two benchmarks. (a,b) depict the progressive improvement in task performance across reflective learning rounds on HLE and GAIA. (c,d) depict the corresponding growth of the skill memory, while organising learned skills into semantically meaningful clusters. }
    \label{fig:placeholder}
\end{figure}

\newpage



\end{titlepage}


\newpage

\section{The Self-Evolving Agent Problem}

\begin{dialogue}[Monday 9:47am, a startup office. The espresso machine is broken.]

\J~\footnote{Tenured theorist.}:~\stage{arrives carrying a thermos of tea, surveying a wall of red Grafana dashboards} Good morning. I see the agent is still performing at exactly 73\%. Remarkable consistency, really. Like a student who reliably gets a C+.

\vspace{4pt}
\Hstudent~\footnote{Second-year CS PhD student.}:~\stage{spins around in chair, three monitors glowing} I tried throwing more GPUs at it over the weekend. Accuracy went from 73.2\% to 73.4\%. Progress!

\vspace{4pt}
\Senior~\footnote{Senior ML engineer, 12 years in production.}:~\stage{without looking up from terminal} That's within the confidence interval, H. You spent \$400 in compute to learn nothing.

\vspace{4pt}
\Hstudent: But what if we fine-tune it on the tickets it got wrong?

\vspace{4pt}
\J: And how many wrong tickets do you have?

\vspace{4pt}
\Hstudent: \ldots about 200.

\vspace{4pt}
\J: \stage{sips tea} You'd overfit before the loss function finished its first cup of coffee. No. What we need is a system that learns the way \textit{you} learn, H -- by remembering your mistakes and not repeating them. Not by rewriting your neurons.

\vspace{4pt}
\Senior: So, a database.

\vspace{4pt}
\J: \stage{smiling} A very \textit{principled} database. With convergence guarantees.

\vspace{4pt}
\Senior: \stage{finally looks up} You had me at ``database'' and lost me at ``convergence guarantees.'' But fine. Show me the architecture.

\vspace{4pt}
\J: \stage{uncaps a marker, draws a loop on the whiteboard} Read from memory. Act. Get feedback. Write to memory. Repeat. I call it Read--Write Reflective Learning.

\vspace{4pt}
\Hstudent: That's just\ldots\ a for-loop with a vector store.

\vspace{4pt}
\J: \stage{beaming} Exactly! But a for-loop with \textit{convergence guarantees}.

\vspace{4pt}
\Senior: \stage{sighs, opens a new terminal tab} Fine. I'll build it. You prove it. H, you benchmark it. Let's go.

\end{dialogue}

\subsection[\Stag~Why Frozen LLMs Need External Memory]{\Stag\footnote{This is the shared track, which presents material common to both research track and practitioner tracks.}~Why Frozen LLMs Need External Memory}
\label{sec:motivation}

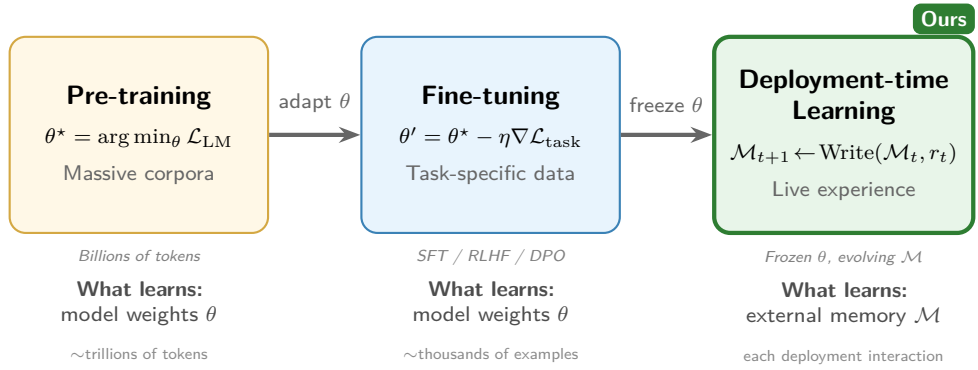
\begin{figure}[t]
\centering
\begin{tikzpicture}[
  font=\sffamily\small,
  phase/.style={draw, rounded corners=6pt, minimum height=2.6cm, minimum width=3.4cm,
    align=center, line width=0.8pt, inner sep=6pt},
  timeline/.style={-{Stealth[length=8pt]}, line width=1.6pt, black!60},
  phaselabel/.style={font=\sffamily\footnotesize\bfseries, above=2pt},
  detail/.style={font=\sffamily\scriptsize, text width=3.0cm, align=center, text=black!70},
  era/.style={font=\sffamily\scriptsize\itshape, below=2pt, text=black!50},
]

\node[phase, fill=theoryBG, draw=theoryBorder] (pt) {
  \textbf{Pre-training}\\[3pt]
  {\scriptsize $\theta^{\star} = \arg\min_\theta \mathcal{L}_{\mathrm{LM}}$}\\[2pt]
  {\scriptsize \textcolor{black!60}{Massive corpora}}
};
\node[era] at (pt.south) {{\fontsize{6}{7}\selectfont Billions of tokens}};

\node[phase, fill=practiceBG, draw=practiceBorder, right=1.2cm of pt] (ft) {
  \textbf{Fine-tuning}\\[3pt]
  {\scriptsize $\theta' = \theta^{\star} - \eta\nabla\mathcal{L}_{\mathrm{task}}$}\\[2pt]
  {\scriptsize \textcolor{black!60}{Task-specific data}}
};
\node[era] at (ft.south) {{\fontsize{6}{7}\selectfont SFT / RLHF / DPO}};

\node[phase, fill=insightBG, draw=insightBorder, line width=1.6pt, right=1.2cm of ft] (dtl) {
  \textbf{Deployment-time}\\
  \textbf{Learning}\\[3pt]
  {\scriptsize $\mathcal{M}_{t+1} \!\leftarrow\! \mathrm{Write}(\mathcal{M}_t, r_t)$}\\[2pt]
  {\scriptsize \textcolor{black!60}{Live experience}}
};
\node[era] at (dtl.south) {{\fontsize{6}{7}\selectfont Frozen $\theta$, evolving $\mathcal{M}$}};

\node[fill=insightBorder, text=white, font=\sffamily\bfseries\scriptsize,
      rounded corners=3pt, inner sep=3pt, above=0pt of dtl.north east, anchor=south east] {Ours};

\draw[timeline] (pt.east) -- (ft.west)
  node[midway, above=4pt, font=\sffamily\scriptsize, text=black!60] {adapt $\theta$};
\draw[timeline] (ft.east) -- (dtl.west)
  node[midway, above=4pt, font=\sffamily\scriptsize, text=black!60] {freeze $\theta$};

\node[detail, below=14pt of pt] (w1) {\textbf{What learns:}\\model weights $\theta$};
\node[detail, below=14pt of ft] (w2) {\textbf{What learns:}\\model weights $\theta$};
\node[detail, below=14pt of dtl] (w3) {\textbf{What learns:}\\external memory $\mathcal{M}$};

\node[font=\sffamily\tiny, text=black!45, below=1pt of w1] {$\sim$trillions of tokens};
\node[font=\sffamily\tiny, text=black!45, below=1pt of w2] {$\sim$thousands of examples};
\node[font=\sffamily\tiny, text=black!45, below=1pt of w3] {each deployment interaction};

\end{tikzpicture}
\caption{The three paradigms of LLM adaptation. \textbf{Pre-training} and \textbf{fine-tuning} update the model parameters $\theta$ and require large data and compute budgets. \textbf{Deployment-time learning} (this work) keeps $\theta$ frozen and instead accumulates experience in an external skill memory $\mathcal{M}$, enabling continual adaptation from live interactions at zero retraining cost.}
\label{fig:paradigms}
\end{figure}

Modern machine learning is about learning from experience~\cite{silver2025welcome,turing2004intelligent}. At the forefront of this evolution, Large Language Models (LLMs) have fundamentally reshaped the learning paradigm, demonstrating exceptional performance across diverse scenarios through few-shot learning~\cite{brown2020language}, supervised fine-tuning~\cite{wei2022finetuned}, and post-training~\cite{guo2025deepseek}. 
Despite their promise, however, achieving practical utility typically requires parameter optimisation via backpropagation~\cite{rumelhart1986learning}, which in turn demands vast amounts of data and computational resources. In practice, the cost and complexity of continual parameter updates mean that most LLM agents are deployed as frozen models~\cite{zhao2023survey}: their parameters $\theta$ remain fixed after pre-training (Figure~\ref{fig:paradigms}). When such an agent encounters a novel task, it draws only on knowledge encoded in $\theta$ and whatever fits in its context window.

\profJ{This is the key premise. If $\theta$ is fixed, all adaptation must come from the \emph{input} -- the prompt, the context, or in our case, the memory. Everything else is just expensive gradient descent cosplay.}

\begin{figure}[t]
\centering
\begin{tikzpicture}[
  node distance=1.2cm and 1.8cm,
  block/.style={draw, rounded corners=4pt, minimum height=1cm, minimum width=2.4cm,
    font=\sffamily\small, align=center, line width=0.8pt},
  arr/.style={-{Stealth[length=6pt]}, thick},
  darr/.style={-{Stealth[length=6pt]}, thick, dashed},
]
  \node[block, fill=practiceBG, draw=practiceBorder] (state) {State $s_t$\\{\scriptsize(New ticket)}};
  \node[block, fill=theoryBG, draw=theoryBorder, right=of state] (read) {READ\\{\scriptsize $c_t \sim \mu(\cdot | s_t, \mathcal{M}_t)$}};
  \node[block, fill=codeBG!20, draw=codeFrame, right=of read] (llm) {\color{black}LLM Act\\{\scriptsize $a_t \sim p_{\mathrm{LLM}}(\cdot | s_t, c_t)$}};
  \node[block, fill=warningBG, draw=warningBorder, below=of llm] (env) {Environment\\{\scriptsize feedback $r_t$, $s_{t+1}$}};
  \node[block, fill=insightBG, draw=insightBorder, below=of read] (write) {WRITE\\{\scriptsize $\mathcal{M}_{t+1} \leftarrow \mathrm{Write}(\mathcal{M}_t, s_t, a_t, r_t)$}};
  \node[block, fill=keyresultBG, draw=keyresultBorder, below=of state] (memory) {Skill\\Memory $\mathcal{M}_t$};

  \draw[arr] (state) -- (read);
  \draw[arr] (read) -- (llm);
  \draw[arr] (llm) -- (env);
  \draw[arr] (env) -- (write);
  \draw[arr] (write) -- (memory);
  \draw[arr] (memory) -- (read);
  \draw[darr, black!40] (memory) -- (state) node[midway,left,font=\scriptsize\sffamily]{next};
\end{tikzpicture}
\caption{Overview of the Read--Write Reflective Learning loop. Given a new task, the agent retrieves a relevant skill from the skill memory (\textsc{Read}), executes it through the frozen LLM (\textsc{Act}), and uses the resulting feedback to reflectively optimise and update the skill library (\textsc{Write}). The LLM parameters remain fixed throughout; all adaptation occurs in the memory.}
\label{fig:loop}
\end{figure}
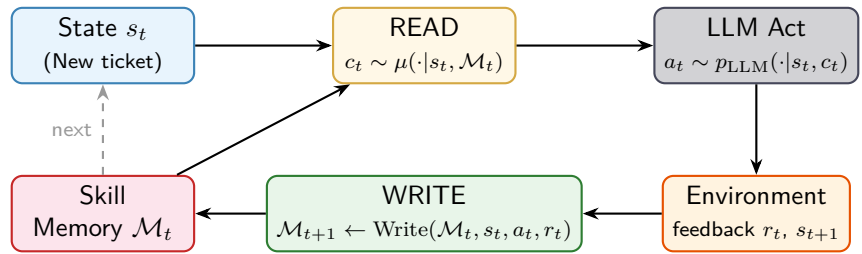

This creates a fundamental limitation: the agent is stateless and it cannot learn from its own deployment experience. The Stateful Reflective Decision Process (SRDP)~\cite{wang2025memento2} resolves this by augmenting the agent with an episodic memory $\mathcal{M}_t$ that grows over time (Figure~\ref{fig:loop}):

\begin{equation}
  \pi^{\mu}(a \mid s, \mathcal{M}_t) = \sum_{c \in \mathcal{M}_t} \mu(c \mid s, \mathcal{M}_t)\; p_{\mathrm{LLM}}(a \mid s, c),
  \label{eq:composite}
\end{equation}
where $p_{\mathrm{LLM}}$ denotes the LLM decision kernel, $s$ is the current state, $c$ represents a retrieved case from the episodic memory $\mathcal{M}_t$, and $\mu$ is the retrieval policy.

\studentH{Wait -- so the LLM doesn't change, but the policy changes because the memory changes? That's like\ldots\ levelling up in a game without upgrading your character. You just get better items.}

\seniorS{I prefer to think of it as a cache that makes you smarter. Which is basically what senior engineers are -- junior engineers with better caches.}

\subsection[\Rtag~Stateful Reflective Decision Process]{\Rtag\footnote{This is research track, which presents theory setup, convergence proofs, and KL-regularised routing analysis.}~Stateful Reflective Decision Process}
\label{sec:formal setup}

\begin{theorytrack}[Formal Setup]

\begin{definition}[Skill Memory]
A skill memory $\mathcal{M}_t = \{c_i\}_{i=1}^{N_t}$ is a finite, growing collection of reusable skill artefacts. Unlike traditional episodic memory that logs raw transitions, each $c_i$ encapsulates a declarative specification, prompts, and executable code. The space of all finite skill memories is denoted $\mathfrak{M}$.
\end{definition}

\begin{definition}[SRDP]
\label{def:srdp}
$\mathcal{D}_{\mathrm{SRDP}} = \langle \mathcal{S}, \mathcal{A}, \mathcal{P}, R, \gamma, \mathfrak{M}, p_{\mathrm{LLM}} \rangle$, extending the standard MDP with episodic memory $\mathfrak{M}$ and an LLM decision kernel $p_{\mathrm{LLM}}(a \mid s, c)$.
\end{definition}

\profJ{The critical insight: by augmenting the state to $x_t := (s_t, \mathcal{M}_t)$, we recover the Markov property. Everything old is new again -- I said this in a 2003 workshop paper.}

The \textbf{Reflected MDP} reformulates this as $\mathcal{D}_{\mathrm{ReMDP}} = \langle \mathcal{X}, \mathcal{C}, \mathcal{P}^{\mathrm{LLM}}, R^{\mathrm{LLM}}, \gamma \rangle$ with transition kernel:
\begin{equation}
  \mathcal{P}^{\mathrm{LLM}}(x' \mid x, c) = \sum_{a \in \mathcal{A}} p_{\mathrm{LLM}}(a \mid s, c)\; \mathbf{1}\{x' = (s', \mathrm{Write}(\mathcal{M}, s, a, r))\}\; \mathcal{P}(s' \mid s, a).
\end{equation}

In Memento-Skills, the $\mathrm{Write}(\mathcal{M}, s, a, r)$ operation is not a simple append. It encapsulates the skill-level reflective updates—performing failure attribution and file-level rewriting to mutate the prompt or code inside $c$. By augmenting the state to $x_t := (s_t,\mathcal{M}_t)$, the system remains Markovian even as the skill library evolves. 

\begin{keyresult}
\begin{theorem}[Convergence, Memento~2~\cite{wang2025memento2}, Thm.~8]
\label{thm:convergence}
Under bounded rewards $|r| \leq R_{\max}$ and $\gamma < 1$, the KL-regularised soft policy iteration over the Reflected MDP converges to the optimal retrieval policy $\mu^*$.
\end{theorem}
\end{keyresult}
\end{theorytrack}


\subsection[\Ptag~From Zero to Self-Evolving Agent]{\Ptag\footnote{This is practitioner track, which presents installation, API, retrieval pipeline, and benchmark recipes.}~From Zero to Self-Evolving Agent}
\label{sec:quickstart}

\begin{practicetrack}[Getting Started in $5$ Minutes]

\studentH{Can I pip-install convergence guarantees?}

\seniorS{No, but you can pip-install the system that has them.}

\vspace{4pt}

\textbf{Installation:}
\begin{lstlisting}[style=darkbash]
git clone https://github.com/Memento-Teams/Memento-Skills.git
cd Memento-Skills
python -m venv .venv && source .venv/bin/activate
pip install -e .          
memento agent   
\end{lstlisting}
\vspace{-4pt}
{\small\color{gray}\faLink~~ PDF copying may mangle this command.
Click \href{https://github.com/Memento-Teams/Memento-Skills\#installation}{\textcolor{codeBlue}{\underline{here}}} to copy from the repo.}

\textbf{\color{white}Configuration} (\texttt{~/memento\_s/config.json}):
\begin{lstlisting}[style=darkbash]
# Memento-S configuration
{
  "llm": {
    "active_profile": "default",
    "profiles": {
      "default": {
        "model": "your-provider/your-model",
        "api_key": "your-api-key",
        "base_url": "https://your-api-url/v1"
      }
    }
  },
  "env": {
    "TAVILY_API_KEY": "your-search-api-key"
  }
}
\end{lstlisting}
\textbf{Your first self-evolving agent}:
\begin{center}
    \includegraphics[width=\linewidth]{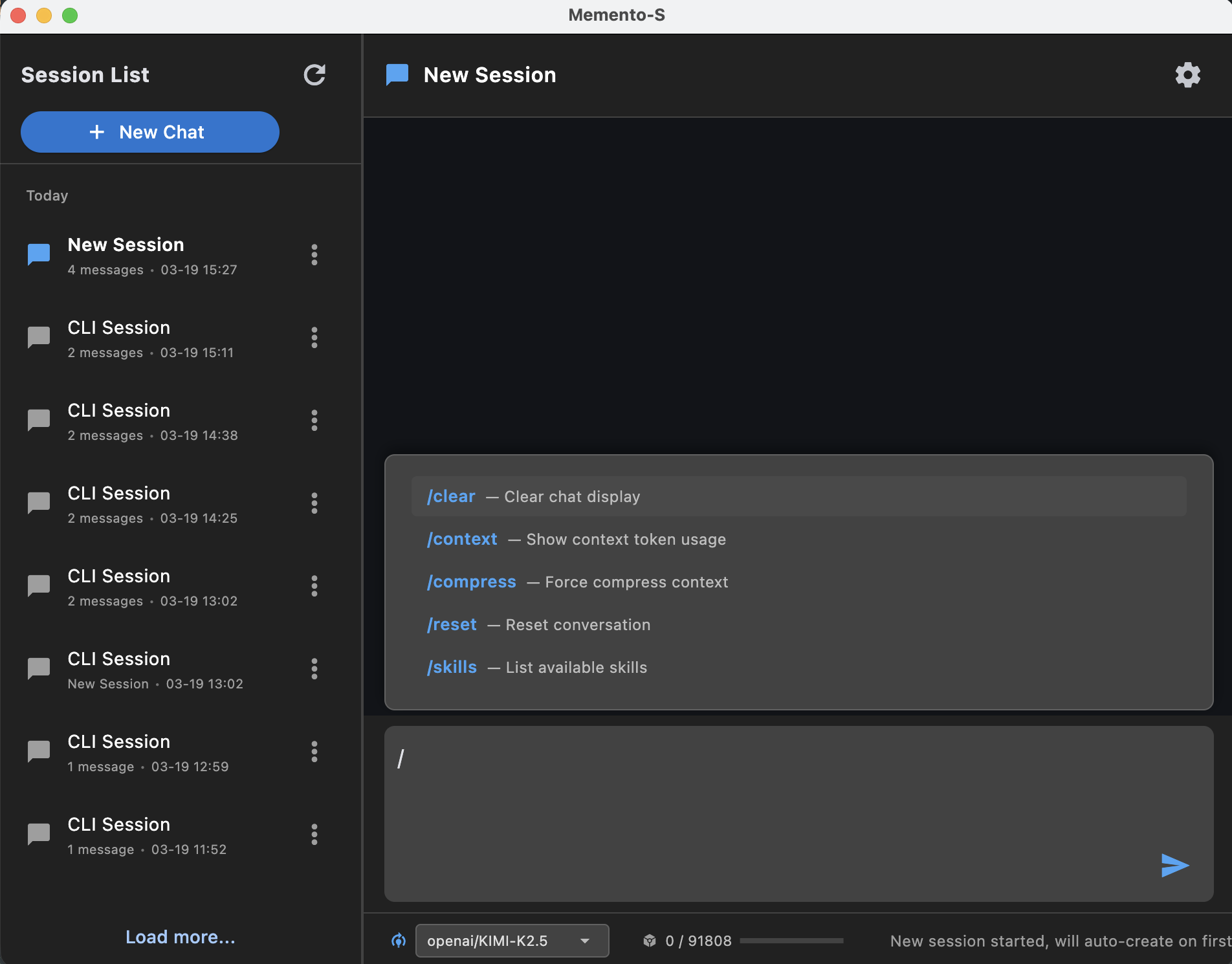}
    \captionof{figure}{The GUI of Memento-Skills.}
    \label{fig:gui}  
\end{center}

\end{practicetrack}

\subsection{\Stag~From Theory to Configuration}
\label{sec:bridge1}
\begin{bridgebox}
In the foundational theory of Memento~2~\cite{wang2025memento2}, Read--Write Reflective Learning is cast as an implicit form of policy iteration operating over raw episodic memory (past states, actions, and rewards). Memento-Skills bridges this theory to production by upgrading the memory unit from passive trajectory logs to an active skill library. Under this skill-centric paradigm, the two key operations take on concrete engineering forms:
\begin{itemize}
    \item \textbf{Writing (Policy Evaluation):} Instead of merely appending interaction logs as in Memento~2, \emph{writing} in Memento-Skills actively mutates the memory. It evaluates execution traces and consolidates the feedback by directly rewriting the reusable skill artefacts (code, prompts, and declarative specs). The policy itself is materialised and optimised within these skill folders.
    \item \textbf{Reading (Policy Improvement):} \emph{Reading} retrieves the most behaviourally relevant skill to guide the frozen LLM. By conditioning the agent's action on an actively refined skill rather than a static prompt or raw historical trace, the system achieves effective policy improvement for the current task.
\end{itemize}

\begin{center}
    \includegraphics[width=\linewidth]{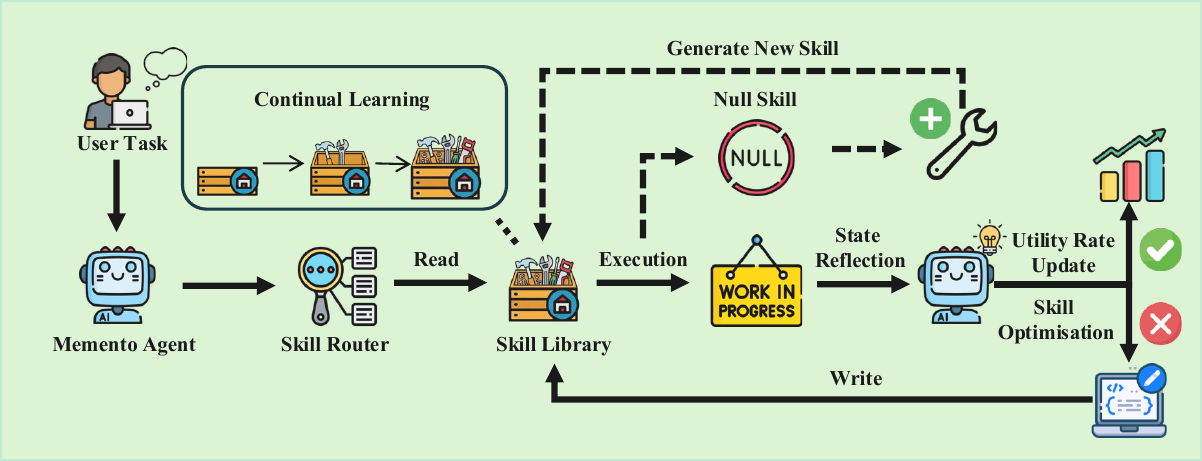}
    \captionof{figure}{The architecture of the Self-Evolving Agent based on Read-Write Reflective Learning. When a user submits a task, the agent uses a skill router to either retrieve an  executable skill from its skill library or generate a new one from scratch, which it then executes to solve the problem. Following execution, the system reflects on the outcome to write back to the library, either by increasing the skill's utility score if the action was successful, or by optimising its underlying skill folders if it failed. This continuous read-write loop enables the agent to progressively expand and refine its capabilities through continual learning, entirely without updating the underlying LLM parameters.}
    \label{fig:framework}
\end{center}

\studentH{Figure~\ref{fig:framework} makes it look so simple. Read a skill, run it, write it back. Kind of elegant, actually.}

\profJ{As it should. A good conceptual figure abstracts away the incidental complexity and reveals the governing loop.}

\seniorS{Sure. And a stick figure abstracts away anatomy. Figure 4 is fine for explaining the idea, but if you want me to trust this thing, I need the engineering drawing --- where the config lives, how the router talks to execution, what gets persisted, and which box wakes me up at 3am.}

\end{bridgebox}

\begin{dialogue}[Tuesday 11am. H has just shared his screen. Everyone wishes he hadn't.]

\stage{An awkward silence. H minimises the figure and opens a file explorer. A single Python file glows ominously on screen: \texttt{main.py} --- 30{,}000 lines.}

\vspace{4pt}
\Hstudent:~\stage{sheepishly} So\ldots\ I do have a working prototype. It's all in one file. But it works!

\vspace{4pt}
\Senior:~\stage{leans in, squints at the screen, then recoils as if physically struck} H. \textit{H.} This is thirty thousand lines in a single file.

\vspace{4pt}
\Hstudent: Twenty-nine thousand, eight hundred and---

\vspace{4pt}
\Senior: \textbf{Do not} finish that sentence. \stage{scrolling furiously} Why is there an \texttt{if-elif} chain from line 4,200 to line 5,700? That's fifteen hundred lines of conditionals. For \textit{what}?

\vspace{4pt}
\Hstudent: Skill routing. Each skill type has its own branch.

\vspace{4pt}
\Senior:~\stage{voice cracking} You hard-coded five thousand \texttt{if-else} clauses. There is a \texttt{elif task\_type == "biology\_question\_about\_frogs"} on line 4,847. \textit{About frogs}, H.

\vspace{4pt}
\Hstudent: Frogs came up a lot in the HLE benchmark\ldots

\vspace{4pt}
\J:~\stage{peering over his glasses with academic detachment} Fascinating. You've essentially hand-compiled a policy table into imperative spaghetti. It's like watching someone implement a hash map with a thousand \texttt{if} statements.

\vspace{4pt}
\Senior:~\stage{has found something worse, whispers} H. Line 12,000 to 18,000 is the skill evolution engine. It's in a function called \texttt{do\_everything\_v3\_final\_FINAL}. There are seven nested \texttt{try-except} blocks, and one of them catches \texttt{BaseException} and just\ldots\ prints ``yolo'' to \texttt{stderr}.

\vspace{4pt}
\Hstudent:~\stage{very quietly} That was a 2am commit.

\vspace{4pt}
\Senior:~\stage{closes laptop, stands up, walks to whiteboard} OK. Here's what's going to happen. \stage{draws a box labelled ``Agent Core'', then six more boxes radiating outward} We are going to take this 30{,}000-line abomination and decompose it. There will be modules. There will be interfaces. There will be \textit{separation of concerns}. And the frog handler is the first thing to go.

\vspace{4pt}
\Hstudent: But it---

\vspace{4pt}
\Senior: \textbf{The frog handler goes.}

\vspace{4pt}
\J:~\stage{quietly amused} You know, S, this is actually a perfect pedagogical example. H built a monolith that \textit{works} but cannot \textit{evolve}. Just like a frozen LLM --- all the knowledge is there, but it's entangled and rigid. The refactoring you're about to do is exactly what the Read--Write loop does to skills: decompose, modularise, and make each unit independently improvable.

\vspace{4pt}
\Senior:~\stage{pauses mid-drawing} \ldots I hate that you just made my code review into a metaphor for your paper.

\vspace{4pt}
\J: Everything is a special case of something I published in 2003.

\end{dialogue}

\begin{bridgebox}

\begin{center}
    \includegraphics[width=\linewidth]{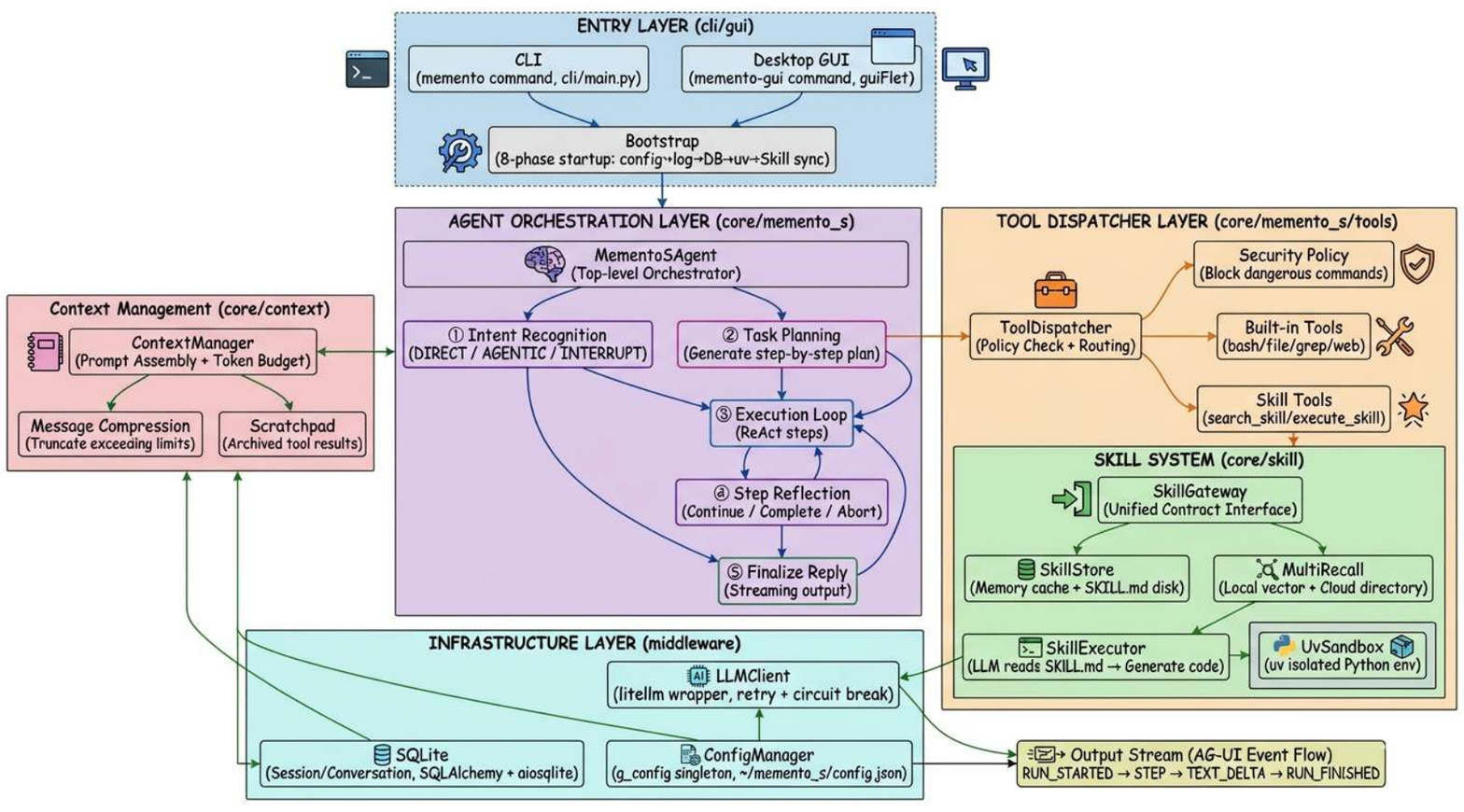}
    \captionof{figure}{Component architecture of Memento-Skills --- or, as S calls it, ``the thing that replaced 30{,}000 lines of \texttt{if-else}.'' The system centres on a Memento-Skills agent that coordinates the LLM client, context manager, built-in tools, and the skills system. The skills system manages both built-in and generated skills, while an evolution engine improves the skill store from task feedback over time.}
    \label{fig:component_framework}
\end{center}

\paragraph{Contributions.} Our main contributions are:
\begin{enumerate}[leftmargin=*,itemsep=2pt]
  \item \textbf{Skill-level reflective learning.} We instantiate the SRDP framework of Memento~2 with a concrete system, Memento-Skills, that treats reusable \emph{skill folders} (code, prompts, and declarative specs) as the unit of memory, enabling continual learning without any parameter updates.
  \item \textbf{Behaviour-aligned skill router.} We train a contrastive retrieval model via single-step offline RL, casting skill routing as a KL-regularised Boltzmann policy that optimises for execution success rather than semantic similarity.
  \item \textbf{Empirical validation.} On GAIA and HLE, Memento-Skills substantially outperforms the static Read-Write baseline, improving test accuracy by 13.7 and 20.8 percentage points, respectively. The results further show that cross-task transfer is strongest when the learned skill library aligns with benchmark domain structure, highlighting when self-evolving skill memory is most effective.
\end{enumerate}
\end{bridgebox}

\newpage
\section{Read--Write Reflective Learning}

\begin{dialogue}[Wednesday 2pm, the whiteboard is covered in dried-out equations]

\J:~\stage{pointing at the learning curve on H's monitor} See that? Accuracy went from 73\% to 84\% in three days. Without touching the model.

\vspace{4pt}
\Hstudent: I plotted the memory coverage radius too. \stage{pulls up a chart} It's decreasing like $O(n^{-1/d})$, just like you predicted!

\vspace{4pt}
\Senior: I'm more interested in \textit{why} it retrieved the wrong case for ticket \#4,721. Customer asked about a refund, agent retrieved a case about password resets. Cosine similarity was 0.91. 

\vspace{4pt}
\J: Ah, the curse of embedding similarity. High cosine doesn't mean \textit{behavioural} utility. In a library of 8,000 skills, semantic overlap is just noise. 

\vspace{4pt}
\Hstudent: Can't we just use end-to-end RL to fine-tune the router? Let the agent learn from its own interaction outcomes?

\vspace{4pt}
\Senior: \stage{deadpan} H, we have 8,000 skills but only a few hundred real-world tasks. The exploration space is a desert. If we wait for the agent to ``stumble'' upon the right skill through random exploration, we'll all be retired before it converges.

\vspace{4pt}
\J: Correct. The exploration-exploitation gap is too wide for on-policy learning. That's why we move to the one-step offline view. We use the LLM as a ``Simulator'' to synthesise a dense field of positive and hard-negative queries. We aren't just matching strings; we are fitting a $Q$-function that predicts execution success before the first token is even generated.

\vspace{4pt}
\Hstudent: \stage{opening a notebook titled ``Things Prof J Says That Turn Out To Be Right''} OK, so synthetic goals, behaviour-aligned routing and then one-step RL. I'm listening.

\end{dialogue}

\subsection{\Stag~The Skill-Level Read--Write Loop}
\label{sec:readwrite-shared}

Memento-Skills is grounded in the theory of Read--Write Reflective Learning~\cite{wang2025memento2}, which provides the theoretical foundation for read--write memory updates as policy iteration. Empirically Memento~\cite{zhou2025memento} and case-based reasoning LLM agents~\cite{guo2024ds,guo2025optimizing} validate this principle across 
deep search, data science, and software engineering. As illustrated in Figure~\ref{fig:framework}, the skill library serves as an external, writable memory, and the agent alternates between (i)~\emph{reading} skills to induce an execution policy for the current goal and (ii)~\emph{writing} updates back to the skill artefacts based on post-hoc reflection. 

This mirrors a policy-iteration view. \emph{Reading} corresponds to policy improvement: the agent retrieves the most relevant skill via a router conditioned on the current query and the accumulated tip memory, then executes the skill's multi-step workflow to produce an answer. \emph{Writing} closes the loop by combining policy evaluation and policy improvement at the skill level: the agent first evaluates by recording execution outcomes and diagnostic traces, then improves by using those traces to revise the skill artefacts that will govern future episodes. Crucially, the memory is not limited to episodic traces but consists of reusable skills, each containing a declarative specification (\texttt{SKILL.md}) together with helper scripts and prompts. Because the write operation rewrites the prompt or program that will be executed next, each write step directly improves the policy embodied in the skill.

This self-evolving mechanism draws on a principle familiar from biological motor learning~\cite{magill2010motor}: early in skill acquisition, performance depends on deliberate, high-level planning; with repeated practice, neural pathways consolidate and execution becomes increasingly automatic~\cite{bacmeister2022motor}. Analogously, a newly created skill in Memento-Skills may be brittle and narrowly scoped, but through iterative revision it is consolidated into a robust, reusable routine, finally forming muscle memory for recurring task patterns. Existing approaches to automatic skill learning either produce text-only guides that amount to prompt optimisation~\cite{agrawal2025gepareflectivepromptevolution,tan2026gskill,mi2026procmem} or overfit to single-task trajectories with limited transferability~\cite{letta2025skilllearning}. 

In contrast, Memento-Skills learns executable, multi-artefact skills and refines them through a reflective read-write learning pipeline. Concretely, after a failed attempt, an LLM-based failure attribution selector first examines the full execution trace and the judge's rationale to identify the single skill most responsible for the error, performing credit assignment at the skill level. Given this diagnosis, a skill rewriter then proposes targeted file-level updates that add guardrails or alternative strategies for the observed failure mode while preserving the skill's generality. When the running utility of a skill (its empirical success rate) drops below a threshold, indicating that in-place patching is insufficient, the system escalates to skill discovery: it either restructures the existing skill folder with a fundamentally different approach or synthesises an entirely new skill, expanding the library to cover novel regions of the task space. To prevent regression, all mutations are guarded by an automatic unit-test gate, a synthetic test case is generated, executed through the updated skill, and scored by the judge~\cite{zheng2023judging}.

\begin{insightbox}
The Read--Write loop is the heartbeat of Memento-Skills. Every interaction follows five steps: \textbf{Observe} $\to$ \textbf{Read} $\to$ \textbf{Act} $\to$ \textbf{Feedback} $\to$ \textbf{Write}.
\end{insightbox}

  \begin{pseudobox}[Read--Write Reflective Learning]
  \begin{algorithmic}[1]
  \REQUIRE Utility threshold $\delta$, minimum samples $n_{\min}$, max feedback rounds $K$
  \STATE Initialise skill library $\mathcal{S}_0 \leftarrow \mathcal{S}_{\mathrm{base}}$, tip memory $\mathcal{T}_0 \leftarrow \varnothing$, utility
  table $U_0(s) \leftarrow 0.5\ \forall s$
  \FOR{$t = 0, 1, 2, \ldots$}
    \STATE \textbf{(1) Observe:} Receive task $q_t$; form augmented input $x_t = (q_t, \mathcal{T}_t)$
    \STATE \textbf{(2) Read} \textcolor{trackR}{[Skill Selection]}:
    \STATE \quad Route: $c_t \leftarrow \mathrm{Router}(x_t,\, \mathcal{S}_t)$
    \STATE \quad \textbf{if} $c_t = \varnothing$ \textbf{and} \textsc{CreateOnMiss} enabled:
    \STATE \qquad $c_t \leftarrow \mathrm{CreateSkill}(x_t)$;\quad $\mathcal{S}_t \leftarrow \mathcal{S}_t \cup \{c_t\}$
    \STATE \textbf{(3) Execute:} Execute multi-step workflow $a_t \leftarrow \mathrm{LLM}(x_t, c_t)$
    \STATE \textbf{(4) Feedback} \textcolor{trackR}{[Judge]}:
    \STATE \quad $r_t \leftarrow \mathrm{Judge}(q_t, a_t, a^{\star}_t)$ 
    \STATE \textbf{(5) Write} \textcolor{trackP}{[Reflective Update]}:
    \STATE \quad \textbf{(5a) Utility update:}\quad $U_{t+1}(c_t) \leftarrow \frac{n_{\mathrm{succ}}(c_t)}{n_{\mathrm{succ}}(c_t) +
  n_{\mathrm{fail}}(c_t)}$
    \STATE \quad \textbf{if} $r_t = \textsc{correct}$: \textbf{continue}
    \STATE \quad \textbf{(5b) Tip memory:}\quad $\mathcal{T}_{t+1} \leftarrow \mathcal{T}_t \cup \{\mathrm{GenericTip}(q_t, a_t, r_t)\}$
    \STATE \quad \textbf{(5c) Skill evolution:}
    \STATE \quad\quad $c^{\dagger} \leftarrow \mathrm{TargetSelector}(\mathrm{trace}_t,\, r_t,\, \mathcal{S}_t^{\mathrm{extra}})$
    \STATE \quad\quad \textbf{if} $U_t(c^{\dagger}) < \delta$ \textbf{and} $n(c^{\dagger}) \geq n_{\min}$: 
    \STATE \qquad\quad $c' \leftarrow \mathrm{DiscoverSkill}(c^{\dagger}, x_t, \mathrm{trace}_t)$;\quad $\mathcal{S}_{t+1} \leftarrow \mathcal{S}_t
  \cup \{c'\}$
    \STATE \quad\quad \textbf{else}: \COMMENT{optimise existing skill in-place}
    \STATE \qquad\quad $\mathcal{S}_{t+1} \leftarrow \mathrm{OptimiseSkill}(c^{\dagger}, x_t, \mathrm{trace}_t,\, \mathcal{S}_t)$
    \STATE \quad\quad \textbf{if} \textsc{UnitTestGate}: validate $\mathcal{S}_{t+1}(c^{\dagger})$; rollback on failure
    \STATE \quad \textbf{(5d) Feedback retry} ($\leq K$ rounds):
    \STATE \quad\quad $a'_t \leftarrow \mathrm{LLM}(x_t, c^{\dagger}_{\mathrm{updated}})$;\quad $r'_t \leftarrow \mathrm{Judge}(q_t, a'_t,
  a^{\star}_t)$
    \STATE \quad\quad \textbf{if} $r'_t = \textsc{incorrect}$: \textbf{repeat} (5b)--(5d)
  \ENDFOR
  \end{algorithmic}
  \end{pseudobox}

\profJ{Steps 2 and 5 are exactly policy improvement and policy evaluation. This is not a metaphor -- it is a mathematical identity. I will die on this hill.}

\seniorS{And steps 1--4 are basically what every web server does: receive request, look up cache, generate response, log result. We've been doing ``reflective learning'' in production for decades. We just didn't have convergence guarantees.}

\subsection{\Ptag~Self-evolving Architecture}

\begin{practicetrack}[The core of the self-evolving mechanism]

\begin{center}
    \includegraphics[width=\linewidth]{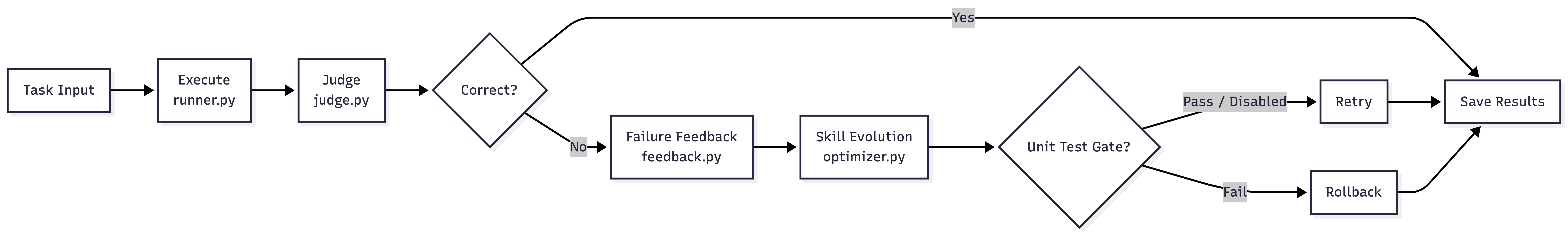}
    \captionof{figure}{This flowchart illustrates the Self-Evolution Engine designed to transform task failures into system growth. It depicts a closed-loop pipeline where an orchestrator audits execution logs to generate, validate, and optimise new skills before persisting them into the global skill catalog.}
    \label{fig:evo_component_framework}
\end{center}

\end{practicetrack}

\subsection{\Rtag~InfoNCE Routing as a One-Step Soft Policy}
\label{sec:infonce_router_theory_short}

\begin{theorytrack}[Contrastive Retrieval as KL-Regularised One-Step RL]

\paragraph{Offline RL Router for Behaviour-Similar Retrieval.}
We find that purely semantic routers (e.g., BM25~\cite{robertson2009probabilistic} or embedding routers such as Qwen-Embedding~\cite{qwen3embedding}) are insufficient for skill selection, because they primarily capture semantic similarity between the user goal and skill text rather than \textbf{behavioural} similarity—i.e., whether executing a skill would produce the desired trajectory and outcome.
To better align routing with execution behaviour, we train the router with single-step offline RL on top of an embedding model, so that retrieval optimises for behaviour similarity instead of lexical or semantic proximity.

\paragraph{Skill database and synthetic query generation.}
In order to train a behaviour-similar retrieval model, we first crawl a local skill database of roughly $8\text{k}$ skills, and randomly sample about $3\text{k}$ skills as seed data to synthesise realistic user routing goals. To align the synthesised goals with the agent's logic stream, we generate queries using only the skill \emph{name} and \emph{description} (without access to the full skill file), and then apply an LLM-based judge~\cite{zheng2023judging} that \emph{does} read the full skill file to filter and verify the quality of the synthetic queries.
This produces high-quality paired data consisting of positive queries (the target skill should be selected) and hard negatives (same domain and terminology, but the target skill is not the right tool). We include the full prompt used for query synthesis in Appendix~\ref{app:router_prompt}.

\paragraph{Router score and multi-positive InfoNCE.}
Let $\mathrm{enc}_\theta(\cdot)$ map a skill document $d$ and a routing goal $q$ to embeddings in $\mathbb{R}^m$:
\[
\mathbf{e}(d)=\mathrm{enc}_\theta(d),\qquad \mathbf{u}(q)=\mathrm{enc}_\theta(q),\qquad
s(d,q)=\mathbf{e}(d)^\top \mathbf{u}(q).
\]
In a minibatch of $B$ skills $\{d_i\}$, each $d_i$ has positives $\mathcal{Q}_i^+$ and hard negatives $\mathcal{Q}_i^-$. Using all in-batch queries
\[
\mathcal{Q}=\bigcup_{k=1}^B(\mathcal{Q}_k^+\cup\mathcal{Q}_k^-),
\]
we minimise the multi-positive InfoNCE loss (temperature $\tau$):
\[
\mathcal{L}_i
=
-\log
\frac{
\sum_{q\in \mathcal{Q}_i^{+}}\exp\left(s(d_i,q)/\tau\right)
}{
\sum_{q\in \mathcal{Q}}\exp\left(s(d_i,q)/\tau\right)
},
\qquad
\mathcal{L}=\frac{1}{B}\sum_{i=1}^{B}\mathcal{L}_i .
\]

\paragraph{One-step offline $Q$-learning view.}
Cast routing as a one-step MDP: state $q$, action $d$, reward $r(q,d)$ indicating whether $d$ is the right skill. With horizon $1$,
\[
Q^\star(q,d)=\E[r(q,d)].
\]
We interpret the learned score as a soft $Q$-function, $Q_\theta(q,d)\propto s(d,q)$, yielding a Boltzmann routing policy
\[
\pi_\theta(d\mid q)=\frac{\exp(Q_\theta(q,d)/\tau)}{\sum_{d'}\exp(Q_\theta(q,d')/\tau)}.
\]
This policy is equivalently the maximiser of a KL-regularised objective (uniform prior $\pi_0$):
\[
\pi^*(\cdot\mid q)=\arg\max_{\pi}\Big\{\E_{d\sim\pi}[Q_\theta(q,d)]-\tau\,\mathrm{KL}(\pi\,\|\,\pi_0)\Big\}.
\]

\studentH{So a small $\tau$ means ``I'm pretty sure—pick this one,'' while a large $\tau$ means ``no rush—spread probability mass around and take a broader look.''}

\paragraph{Why InfoNCE matches ``policy fitting'' in one step.}
InfoNCE has the form ``push up positives, push down competitors'' under the same softmax normaliser used by $\pi_\theta$. Hence minimising $\mathcal{L}$ is (approximately, via in-batch normalisation) maximum-likelihood training that makes $\pi_\theta$ place high probability mass on the logged rewarding pairs (positives) while suppressing hard negatives—i.e., single-step offline policy improvement for routing.

\end{theorytrack}

\subsection{\Ptag~Implementing the Retrieval Pipeline}
\label{sec:retrieval-code}

\begin{practicetrack}[The Retrieval Engine]

\seniorS{Here's the core retrieval class. Every line maps to an equation. I added the references in comments so H stops asking ``but why?''}

\begin{center}
    \includegraphics[width=\linewidth]{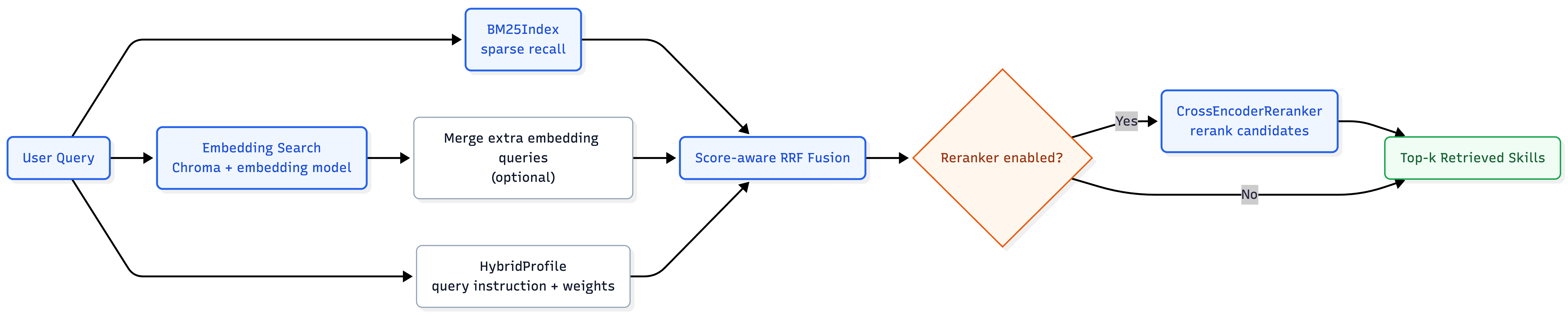}
    \captionof{figure}{Overview of the retrieval pipeline in Memento-Skills. The system combines sparse BM25 recall and dense embedding-based retrieval, fuses the candidates with score-aware reciprocal rank fusion, and optionally applies a cross-encoder reranker to produce the final top-k skills.}
    \label{fig:retrieval_component_framework}
\end{center}

\end{practicetrack}

\begin{sharedtrack}[Router Evaluation]

\paragraph{Skill source filtering and deduplication.}
We first collect candidate skills from public GitHub repositories and unify them into a JSONL catalog. To retain only mature and broadly adopted skills, we keep entries with \texttt{stars} $>500$ and drop the rest. We then normalise description whitespace, compute a SHA-256 hash of each normalised description, and deduplicate by hash to remove duplicated or near-duplicated skills. When multiple rows share the same hash, we keep a single representative by a deterministic score: higher \texttt{stars}, then newer \texttt{updatedAt}, then lexicographically larger \texttt{id}. We optionally apply a second pass of name-level deduplication with the same tie-breaking rule. The resulting curated catalog is used as the base skill universe for router training data generation. We publicly open-source the dataset at~\url{https://skills.memento.run/market/}.

\begin{center}    \includegraphics[width=\linewidth]{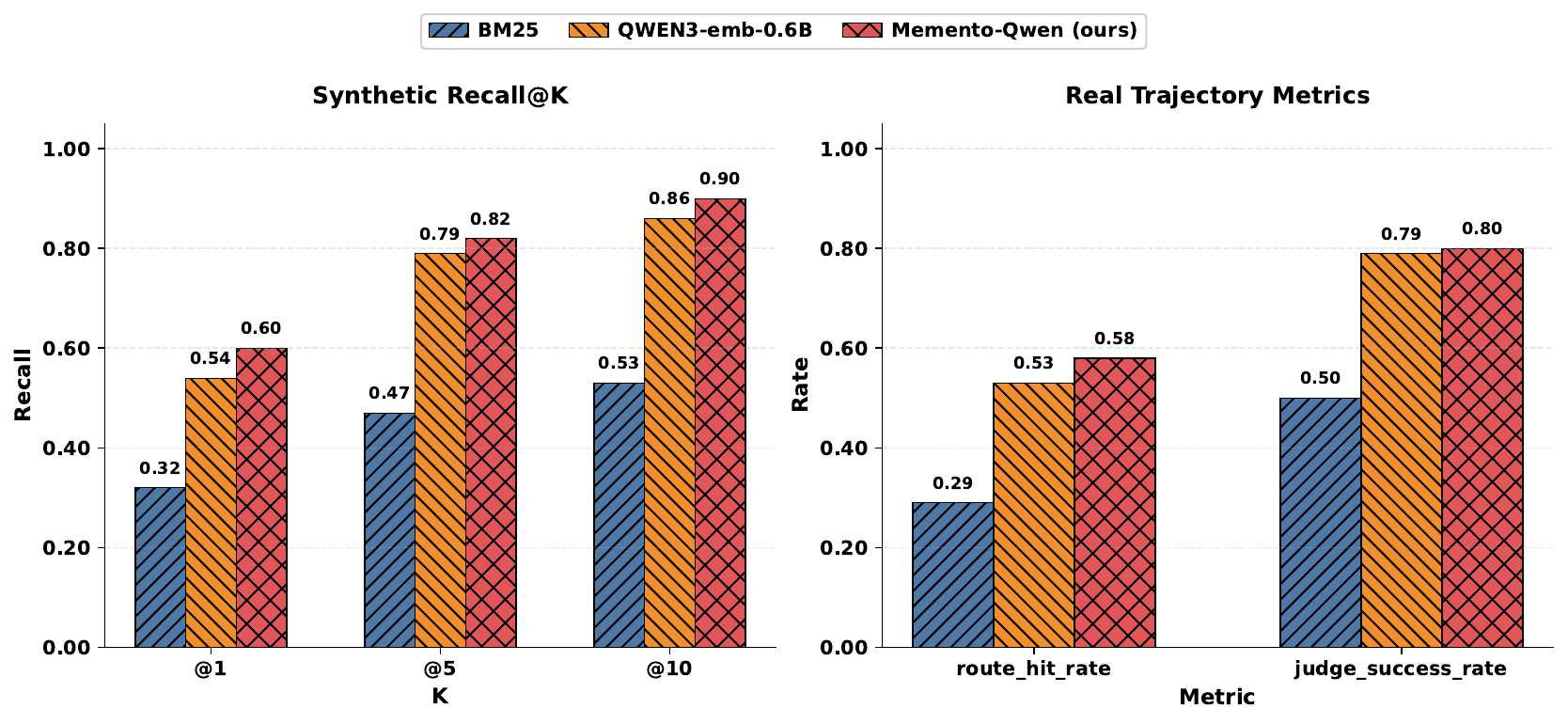}
\captionof{figure}{Router performance evaluation. Left: Offline recall of three routing models evaluated on synthetic query-skill pairs. Right: End-to-end execution success rates for each router.}
\label{fig:router}
\end{center}

\paragraph{Experimental Setting.} We evaluate the performance of the skill router from two complementary angles: (i) offline retrieval quality on synthetic queries, and (ii) end-to-end effectiveness on real execution trajectories. We use the Qwen3-Embedding-0.6B~\footnote{\url{https://huggingface.co/Qwen/Qwen3-Embedding-0.6B}} as embedding model.

\paragraph{Results.} We report Recall@K over $140$ synthetic routing queries, where a query is a hit if the ground-truth skill appears in the top-K candidates. As shown in Fig.~\ref{fig:router} (left), Memento-Qwen consistently outperforms both BM25 and the Qwen3 embedding baseline across all values of K. Most notably, Recall@1 rises from $0.32$ (BM25) and $0.54$ (Qwen3) to $0.60$, a relative gain of $10\%$ over the strongest semantic baseline. By K=10 the gap widens to $0.90$, indicating that behaviour-aligned training not only sharpens the top-1 pick but also populates the candidate list with more relevant alternatives.

To test whether offline retrieval gains translate into real execution improvements, we measure two end-to-end metrics: \emph{route hit rate} (whether the router’s top-1 choice is an appropriate skill for the task) and \emph{judge success rate} (whether the full trajectory actually solves the task). Fig.~\ref{fig:router} (right) reveals that Memento-Qwen lifts route hit rate from $0.29$ (BM25) and $0.53$ (Qwen3) to $0.58$, and judge success rate from $0.50$ and $0.79$ to $0.80$. The disproportionately large improvement over BM25 confirms that lexical matching is a poor proxy for behavioural utility: many skills share domain terminology yet require fundamentally different execution strategies. Meanwhile, the smaller but consistent gain over Qwen3 shows that even dense semantic embeddings under-represent execution-relevant features, and that the single-step RL fine-tuning effectively injects behavioural signal into the embedding space.
\end{sharedtrack}

\newpage
\section{Self-Evolving Evaluation}

\begin{dialogue}[Wednesday 2pm, Zoom call. Cameras on.]

\Hstudent:~\stage{shrugs, sharing screen with a confusion matrix} Look, synthetic data is enough. Generate 10K queries, measure classification accuracy, call it router quality. I can have results by Friday. Paper by Monday.

\vspace{4pt}
\J:~\stage{leans forward, frowning at the matrix} Not enough. Accuracy is a proxy. Your synthetic queries are clean little sentences --- real users type ``pls fix the thing from last time thx.'' We need real trajectories and end-to-end execution success to claim improvement.

\vspace{4pt}
\Hstudent:~\stage{eyes lighting up} Wait --- so you're saying we need a \textit{second} evaluation? That's a second paper. ``On the Insufficiency of Offline Metrics for Skill Routing.'' I can see the title already.

\vspace{4pt}
\Senior:~\stage{deadpan, arms crossed} You are missing the point. If the agent retrieves a case that says ``delete the user's config and start fresh'' and the LLM \textit{executes it}, none of your metrics matter. The customer's environment is on fire and your confusion matrix says 94\%.

\vspace{4pt}
\Hstudent:~\stage{already typing a new LaTeX file} ``Safety-Aware Evaluation of Autonomous Skill Retrieval Systems.'' That's \textit{three} papers, S. You just gave me a third paper.

\vspace{4pt}
\Senior: I gave you a production incident. Please stop turning my trauma into publications.

\vspace{4pt}
\J:~\stage{pointing at the camera} H, focus. ``Looks correct on paper'' and ``works end-to-end'' are fundamentally different failure modes. I have a 2007 paper about this.

\vspace{4pt}
\Hstudent: Can I cite it?

\vspace{4pt}
\J: You \textit{must} cite it.

\vspace{4pt}
\Senior:~\stage{tilts head} And ``works'' and ``safe to run in production'' are different failure modes \textit{again}. I have a 3am incident report about this.

\vspace{4pt}
\Hstudent: Can I cite \textit{that}?

\vspace{4pt}
\Senior: It's in a private Slack channel marked \texttt{\#incident-2023-pain}, so no.

\vspace{4pt}
\Hstudent:~\stage{typing reluctantly, adding rows to a spreadsheet} Fine. So what do we actually put in \textit{this} paper? I'm trying to stay under a page limit here, but every time one of you opens your mouth, I gain a new ablation study.

\vspace{4pt}
\J:~\stage{counting on fingers} Two validations we can run now. One: synthetic retrieval quality --- does the router pick the right case? Two: trajectory success --- does the full loop actually solve the task?

\vspace{4pt}
\Senior:~\stage{nods once} And we state clearly: each one covers a different failure mode. Passing both is necessary. Passing only one is a press release, not a result. Sandbox safety --- whether it solves the task without breaking anything else --- is the third axis, but that requires a proper isolation harness. Future work.

\vspace{4pt}
\Hstudent:~\stage{muttering while typing, a dangerous gleam in his eyes} Three benchmarks. Three failure modes. Three papers. I'm naming them Goku, Vegeta, and Piccolo. Goku is the main paper --- strongest contribution, hits first. Vegeta is the follow-up --- technically impressive, slightly bitter about being second. Piccolo is the safety paper --- everyone forgets about it, but it saves the day in the end.

\vspace{4pt}
\Senior: We need one paper, H. \textit{One.}

\vspace{4pt}
\Hstudent: Goku can fuse with Vegeta. That's canon.
\end{dialogue}


\subsection{\Stag~Experimental Setup and Results}
\label{sec:benchmarks}

\begin{sharedtrack}[Which Benchmark is suitable for Memento-Skills?]

\paragraph{Experimental Settings.} To validate the progressive capability expansion and skill-learning proficiency of Memento-Skills, we evaluate our system on two representative benchmarks: General AI Assistants (GAIA)~\cite{mialon2023gaia} and Humanity's Last Exam (HLE)~\cite{phan2025humanitysexam}. These datasets naturally align with our objective of testing an agent's ability to create, refine, and reuse skills across diverse reasoning tasks.

\paragraph{General AI Assistants (GAIA).} 
GAIA~\cite{mialon2023gaia} comprises non-trivial, real-world questions with unambiguous answers that demand a combination of multi-step reasoning, multi-modality handling, web browsing, and general tool use. This environment serves as an ideal testbed for our skill-learning scenario, requiring the agent to dynamically synthesise and apply distinct skills to solve varied problems. From the GAIA validation set, we utilise 165 questions, splitting them into 100 training examples and 65 test examples.

\paragraph{Humanity's Last Exam (HLE).} Developed by human experts, HLE~\cite{phan2025humanitysexam} is designed to assess the limits of broad-domain reasoning and contains 2,500 questions across 8 diverse academic subjects (e.g., mathematics, humanities, and natural sciences). For our experiments, we sample a subset of questions evenly distributed across these categories, resulting in 788 training examples and 342 test examples. This structure allows us to evaluate how effectively Memento-Skills leverages and transfers learned skills between different questions within the same subject domain.

\paragraph{Baselines.} To isolate the contribution of the self-evolving mechanism, we compare Memento-Skills (the full system) against a Read-Write ablation that retains the same read--write reflective learning loop---skill retrieval, LLM execution, and feedback collection---but disables all skill-level optimisation: no failure attribution, no skill rewriting, and no skill discovery. All the experiments in this paper use the Gemini-3.1-Flash as the underlying LLM.

\end{sharedtrack}

\begin{sharedtrack}[Results of GAIA.]

We evaluate Memento-Skills on the GAIA benchmark with a maximum of three reflective retries per question. As shown in Figure~\ref{fig:gaia}, the self-evolving mechanism continuously refines the skill library through iterative interaction: the overall training success rate climbs from $65.1\%$ on the first attempt to $91.6\%$ by the third round. On the unseen test set, the full Memento-Skills system achieves $66.0\%$ overall accuracy, compared with $52.3\%$ for the Read-Write ablation, confirming that the skill optimisation pipeline contributes a $13.7$ percentage-point gain beyond what retrieval and execution alone can provide.

\paragraph{Limited cross-task transfer on GAIA.} The gap between training-peak and test-set accuracy reveals an important structural property of the benchmark: GAIA questions are highly diverse, with little overlap in the reasoning patterns required. A case study confirmed that most skills optimised during training were never triggered during testing, because no sufficiently similar test question existed. This result suggests that skill transfer depends on domain alignment, a hypothesis we test directly on HLE below, where structured subject categories provide natural opportunities for reuse.

\begin{center}
  \includegraphics[width=\linewidth]{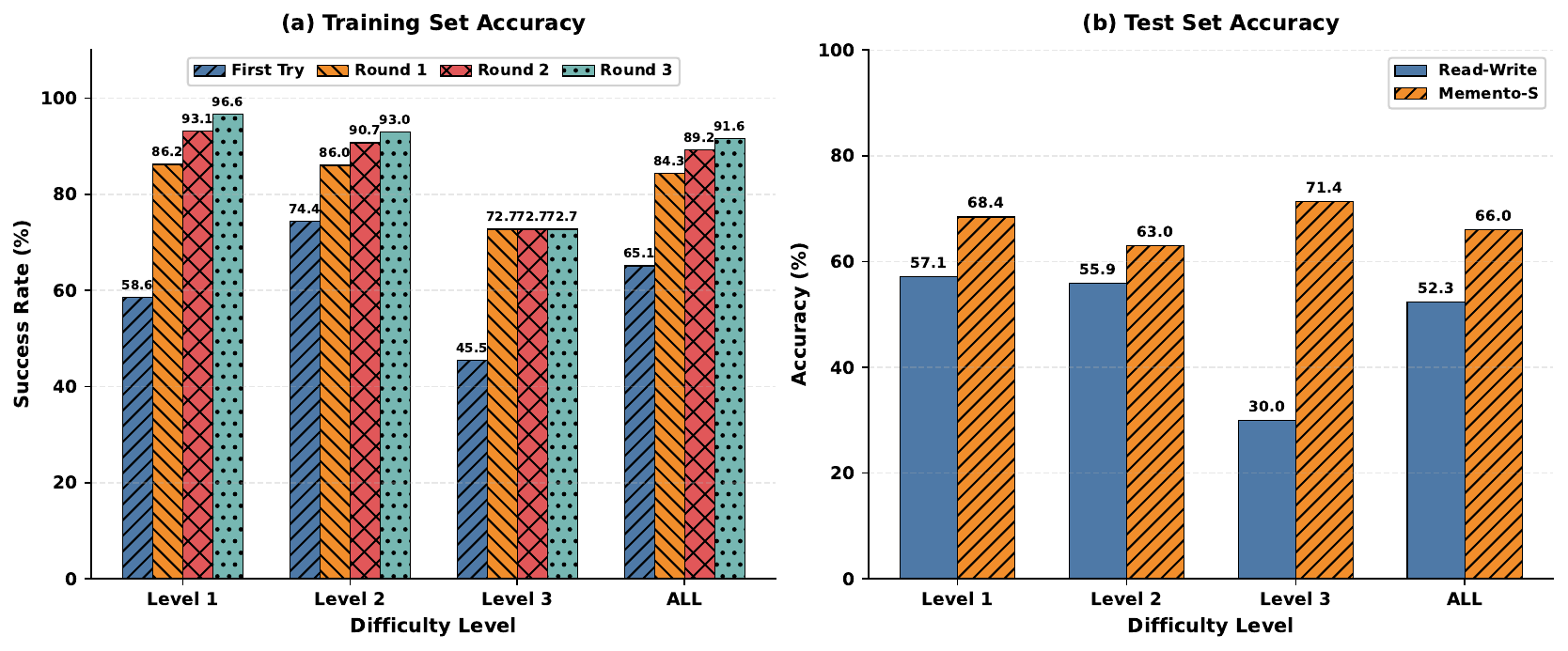}
  \captionof{figure}{GAIA results: training accuracy across retries (left) and test-set comparison with the Read-Write baseline (right).}
  \label{fig:gaia}
\end{center}

\end{sharedtrack}

\begin{sharedtrack}[Results of HLE.]

Figure~\ref{fig:hle} reports per-category accuracy on HLE across four training rounds (R0--R3) and the final test-set evaluation. During training, the overall success rate rises steadily from $30.8\%$ (R0) to $54.5\%$ (R3), with every subject category showing consistent improvement. Humanities and Biology benefit the most, reaching $66.7\%$ and $60.7\%$ respectively by R3, while Engineering saturates earlier at $42.1\%$, suggesting that some domains are harder to cover with skill-level abstractions alone.

On the test set, Memento-Skills achieves $38.7\%$ overall, more than doubling the Read-Write baseline ($17.9\%$). Unlike GAIA, the structured subject taxonomy of HLE enables substantial skill transfer: a skill refined on one Biology training question is frequently reused for unseen Biology questions in the test set. This confirms that domain-aligned skill libraries are the key enabler of cross-task generalisation.

\begin{center}
  \includegraphics[width=\linewidth]{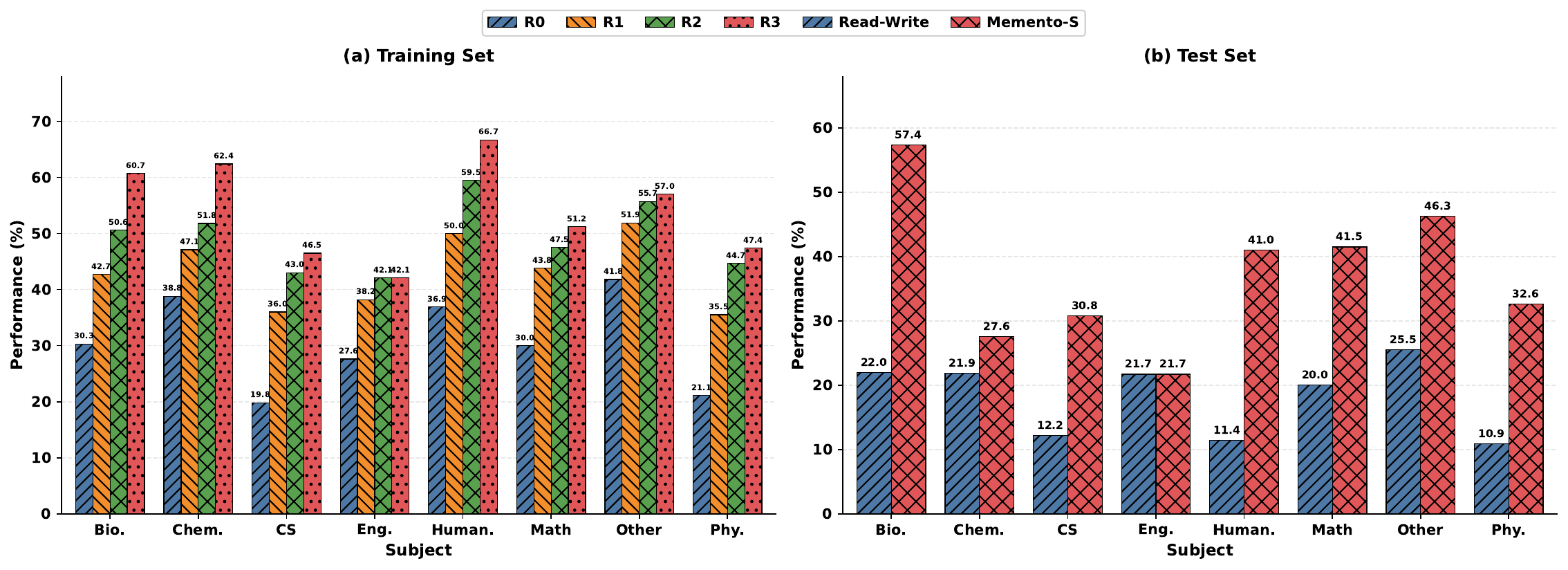}
  \captionof{figure}{HLE results: training accuracy across retries (left) and test-set comparison with the Read-Write baseline (right).}
  \label{fig:hle}
\end{center}

\end{sharedtrack}




\begin{sharedtrack}[Skill Library Growth.]

Figure~\ref{fig:skill_scatter} visualises the skill library after learning on each benchmark via t-SNE projections of skill embeddings. Starting from the same 5 atomic skills (red stars), GAIA learning produces a compact library of 41 skills, reflecting the benchmark's diverse but relatively small question set. In contrast, HLE learning expands the library to 235 skills that spread across a much wider embedding space, mirroring the breadth of its 8 academic domains. Notably, the learned skills (blue dots) cluster into semantically coherent neighbourhoods; each cluster corresponds to a domain-specific capability the agent acquired through reflective self-evolution. This progressive densification of the embedding space is precisely the mechanism that drives the convergence phenomenon analysed in the Bridge below: as the library grows denser, the memory coverage radius $r_\mathcal{M}$ shrinks, and the performance gap narrows.

\begin{center}
  \includegraphics[width=\linewidth]{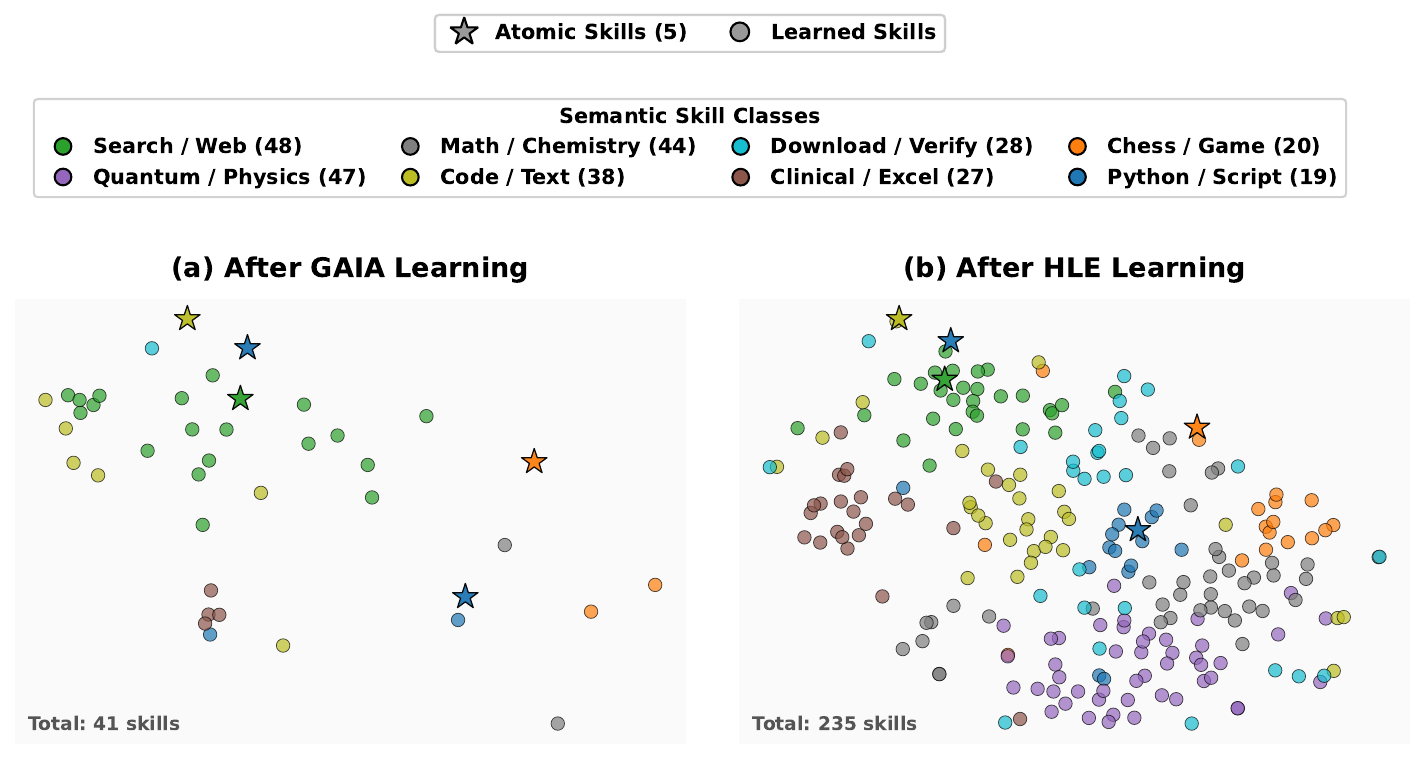}
  \captionof{figure}{t-SNE projection of skill embeddings. Red stars denote the 5 atomic (seed) skills; blue dots denote skills learned through reflective self-evolution. (a)~After GAIA learning the library grows to 41 skills. (b)~After HLE learning the library expands to 235 skills spanning diverse academic domains.}
  \label{fig:skill_scatter}
\end{center}

\end{sharedtrack}

\subsection{\Stag~From LLM Competence Radius to Embedding Quality}
\label{sec:bridge2}

\begin{bridgebox}

\sffamily\small
Look again at Figure~\ref{fig:hle}: training accuracy climbs from $30.8\%$ (R0) to $54.5\%$ (R3), with the steepest gain in the first round and progressively smaller increments thereafter.
Two things are happening simultaneously with each round: \textbf{(i)}~existing skills are refined: reflection patches failure modes, so each skill covers a wider neighbourhood of queries; and \textbf{(ii)}~new skills are added to the library, shrinking the gaps between covered regions.
Together, these two forces drive the \emph{diminishing-returns} curve we observe: early rounds yield large jumps because the library is sparse and skills are rough; later rounds yield smaller gains because most of the reachable space is already well-covered.
Figure~\ref{fig:skill_scatter} makes this concrete: comparing the GAIA library (41 skills) with the HLE library (235 skills), we see that additional learning rounds fill in the gaps between existing clusters until the embedding space is densely covered, at which point adding more skills yields diminishing returns because nearby skills already exist.

\vspace{4pt}
This convergence behaviour is not accidental; it is exactly what the theory of Memento~2 predicts.
The asymptotic value gap (Corollary~15, Memento~2) decomposes as:
\[
  \underbrace{\sup_s |V^{\pi^*}(s) - V^{\pi_\mathcal{M}}(s)|}_{\text{performance gap}} \leq \frac{2R_{\max}}{(1-\gamma)^2} \bigl(\underbrace{\varepsilon_{\mathrm{LLM}}(r_\mathcal{M})}_{\text{LLM quality}} + \underbrace{\delta_\mathcal{M}}_{\text{retrieval error}}\bigr).
\]
As the library grows (more episodes), the memory coverage radius $r_\mathcal{M}$ shrinks, which simultaneously reduces $\varepsilon_{\mathrm{LLM}}(r_\mathcal{M})$ (the LLM only needs to generalise over a smaller neighbourhood) and $\delta_\mathcal{M}$ (the router is more likely to find a behaviourally relevant skill).  Once both terms are small enough, further rounds produce only marginal improvement: the system has converged.

\vspace{4pt}
The bound also reveals three \emph{independent} knobs for reducing this gap:

\begin{center}
\begin{tikzpicture}[
  knob/.style={draw, circle, minimum size=1.4cm, font=\sffamily\scriptsize, align=center, line width=1pt},
]
  \node[knob, fill=theoryBG, draw=theoryBorder] (k1) at (0,0) {Stronger\\LLM};
  \node[knob, fill=practiceBG, draw=practiceBorder] (k2) at (3.5,0) {More\\Episodes};
  \node[knob, fill=insightBG, draw=insightBorder] (k3) at (7,0) {Better\\Embedding};

  \node[below=0.4cm of k1, font=\sffamily\scriptsize] {reduces $\varepsilon_{\mathrm{LLM}}$};
  \node[below=0.4cm of k2, font=\sffamily\scriptsize] {reduces $r_\mathcal{M}$};
  \node[below=0.4cm of k3, font=\sffamily\scriptsize] {reduces $\delta_\mathcal{M}$};

  \node[above=0.5cm of k2, font=\sffamily\small\bfseries] {Three Independent Knobs};
\end{tikzpicture}
\end{center}

\profJ{The convergence you see in the table is the bound tightening in real time. Each round shrinks $r_\mathcal{M}$, which pulls down both other terms. And because the three knobs are \emph{independent}, you can improve any one without touching the others. That's why the system is modular.}

\seniorS{Translation: the diminishing returns aren't a bug; they're a sign the system is converging. And if I want to push accuracy further, I have three independent levers: upgrade the embedding model on Tuesday, swap in a better LLM on Wednesday, and run more episodes on Thursday.}

\end{bridgebox}

\begin{epilogue}[Friday 5:30pm. The espresso machine has been fixed.]

\Senior:~\stage{showing Grafana} 93.5\% accuracy. p99 latency 195ms. Zero gradient updates. I'm buying the espresso machine a thank-you card.

\vspace{4pt}
\Hstudent: I ran the ablation study. Removing the skill optimisation drops accuracy by 8\%. Removing the Memento-Qwen causes retrieval collapse. The theory\ldots\ actually predicted all of this.

\vspace{4pt}
\J:~\stage{sipping espresso, looking insufferably pleased} I believe the phrase you're looking for is ``Prof J was right.''

\vspace{4pt}
\Senior: Don't push it. But I do want to know: what happens when we hit a million cases? Does the Parzen kernel scale?

\vspace{4pt}
\Hstudent: And can we get the convergence rate? Not just ``it converges'' but ``it converges in $O(n^{-1/d})$ episodes''?

\vspace{4pt}
\J:~\stage{standing, reaching for the whiteboard marker} Those are exactly the right questions. Chapter 3.

\vspace{4pt}
\Senior: \stage{to H, whispering} He planned this. He always plans this.

\end{epilogue}

\newpage

\section{Conclusion}

We have presented Memento-Skills, a system that bridges the gap between memory-based learning and skill-based learning for LLM agents. The central insight is to treat executable skills as the unit of external memory, thereby transferring the theoretical guarantees of the Stateful Reflective Decision Process into a concrete, deployable artefact. Through the Read--Write Reflective Learning loop, the agent autonomously acquires, refines, and reuses these skills from deployment experience alone, requiring no parameter updates to the underlying LLM. A behaviour-aligned contrastive router, trained via single-step offline RL, ensures that retrieval optimises for execution success rather than surface-level similarity. Experiments on GAIA and HLE confirm that this skill-as-memory formulation substantially outperforms a static-library ablation, and that cross-task transfer is strongest when skills are aligned with structured domain categories. More broadly, Memento-Skills demonstrates that continual learning need not reside in model weights: an ever-growing, self-improving skill library can serve as a persistent, non-parametric intelligence layer that any frozen LLM can draw upon.

\section*{Contributions}

\subsection*{Algorithm Team}

Huichi Zhou, University College London \\
Siyuan Guo, Jilin University \\
Anjie Liu, Hong Kong University of Science and Technology (Guangzhou)  \\
Zhongwei Yu, Hong Kong University of Science and Technology (Guangzhou)  \\
Ziqin Gong, Hong Kong University of Science and Technology (Guangzhou)  \\
Bowen Zhao, Hong Kong University of Science and Technology (Guangzhou)  \\
Zhixun Chen, Hong Kong University of Science and Technology (Guangzhou)  \\
Menglong Zhang, Hong Kong University of Science and Technology (Guangzhou)  \\
Yihang Chen, University College London

\subsection*{Engineering Team}

Jinsong Li, AI Lab, The Yangtze River Delta \\
Runyu Yang, AI Lab, The Yangtze River Delta \\
Qiangbin Liu, AI Lab, The Yangtze River Delta \\
Xinlei Yu, AI Lab, The Yangtze River Delta \\
Jianmin Zhou, AI Lab, The Yangtze River Delta \\
Na Wang, AI Lab, The Yangtze River Delta \\
Chunyang Sun, AI Lab, The Yangtze River Delta 

\subsection*{Advisor}

Jun Wang, University College London



{\small\bibliographystyle{plainnat}\bibliography{refs}}

\newpage 
\appendix

\section{Reading Path}

\begin{tcolorbox}[
  enhanced, colback=white, colframe=black!30, boxrule=1pt,
  width=0.92\textwidth, arc=6pt, shadow={2pt}{-2pt}{0pt}{black!10}
]
\sffamily\small
This paper is organised as \textbf{interleaving tracks} for \textbf{two audiences}.
Each section opens with a shared Dialogue, then forks into a
\colorbox{theoryBG}{\color{trackR}\textbf{Research}} track and a
\colorbox{practiceBG}{\color{trackP}\textbf{Practitioner}} track,
before merging at a Bridge.
Pick the path that matches your goal; read both for the complete picture.

\vspace{8pt}
\centering
\begin{tikzpicture}[
  node distance=0.4cm and 0.45cm,
  every node/.style={font=\sffamily\scriptsize},
  rnode/.style={draw=theoryBorder,fill=theoryBG,rounded corners=3pt,
    minimum width=1.4cm,minimum height=0.6cm},
  pnode/.style={draw=practiceBorder,fill=practiceBG,rounded corners=3pt,
    minimum width=1.4cm,minimum height=0.6cm},
  snode/.style={draw=black!30,fill=sharedBG,rounded corners=3pt,
    minimum width=1.4cm,minimum height=0.6cm},
  dnode/.style={draw=dialogueBorder,fill=dialogueBG,rounded corners=3pt,
    minimum width=1.4cm,minimum height=0.6cm},
  bnode/.style={draw=bridgeBorder,fill=bridgeBG,rounded corners=3pt,
    minimum width=1.4cm,minimum height=0.6cm},
  arr/.style={-{Stealth[length=4pt]},thick,black!50},
  lbl/.style={font=\sffamily\tiny,black!55}
]
  \node[dnode] (d1) {Dialogue};
  \node[snode,right=of d1] (s1) {Shared};
  \node[rnode,above right=0.35cm and 0.55cm of s1] (r1) {Theory};
  \node[pnode,below right=0.35cm and 0.55cm of s1] (p1) {Practice};
  \node[bnode,below right=0.35cm and 0.55cm of r1] (b1) {Bridge};

  \draw[arr] (d1)--(s1);
  \draw[arr] (s1)--(r1);
  \draw[arr] (s1)--(p1);
  \draw[arr] (r1)--(b1);
  \draw[arr] (p1)--(b1);

  \node[dnode,right=0.55cm of b1] (d2) {Dialogue};
  \node[snode,right=of d2] (s2) {\ldots};
  \draw[arr] (b1)--(d2);
  \draw[arr] (d2)--(s2);

  \node[lbl,above=0.08cm of r1] {proofs \& convergence};
  \node[lbl,below=0.08cm of p1] {code \& deployment};

\end{tikzpicture}

\vspace{10pt}
\raggedright
\begin{minipage}[t]{0.47\textwidth}
  \colorbox{theoryBG}{\color{trackR}\textbf{\,\faFlask~Researcher Path\,}}\\[4pt]
  {\footnotesize Formal SRDP setup, convergence proofs, and KL-regularised routing analysis.
  Follow the \textbf{cream-shaded} sections.\\[2pt]
  \textit{Dialogue\,$\to$\,Shared\,$\to$\,\textbf{Theory}\,$\to$\,Bridge\,$\to$\,\ldots}}
\end{minipage}
\hfill
\begin{minipage}[t]{0.47\textwidth}
  \colorbox{practiceBG}{\color{trackP}\textbf{\,\faCode~Practitioner Path\,}}\\[4pt]
  {\footnotesize Installation, API walkthrough, retrieval pipeline, and benchmark recipes.
  Follow the \textbf{blue-shaded} sections.\\[2pt]
  \textit{Dialogue\,$\to$\,Shared\,$\to$\,\textbf{Practice}\,$\to$\,Bridge\,$\to$\,\ldots}}
\end{minipage}

\vspace{8pt}
{\footnotesize
\colorbox{dialogueBG}{\textbf{Dialogue}} opens each section with motivation.\quad
\colorbox{sharedBG}{\textbf{Shared}} presents material common to both tracks.\quad
\colorbox{bridgeBG}{\textbf{Bridge}} connects theoretical results to engineering choices.\quad
\colorbox{epilogueBG}{\textbf{Epilogue}} closes the narrative.\quad
\\ Three characters---\textbf{\color{marginJ}J}, \textbf{\color{marginH}H}, and \textbf{\color{marginS}S}---annotate inline throughout.}
\end{tcolorbox}

\section{Characters}

\vspace{10mm}
\begin{tcolorbox}[
  enhanced, colback=white, colframe=dialogueBorder!50, boxrule=1pt,
  width=0.88\textwidth, arc=6pt
]
\centering\sffamily\small
\begin{tabular}{clp{8.5cm}}
  \toprule
  & \textbf{Character} & \textbf{Perspective \& Personality} \\
  \midrule
  {\color{marginJ}\faGraduationCap} & \textbf{J} &
    Tenured theorist. Writes proofs on napkins. Believes everything is a special case of something he published in 2003.
    ``\textit{But does it converge?}'' \\[6pt]
  {\color{marginH}\faFire} & \textbf{H} &
    Second-year CS PhD student. Runs 47 experiments simultaneously and names them all after anime characters. Thinks every problem needs more GPUs.
    ``\textit{What if we just\ldots\ scale it?}'' \\[6pt]
  {\color{marginS}\faServer} & \textbf{S} &
    Senior ML engineer, 12 years in production. Has been paged at 3am enough times to develop a Pavlovian response to Slack notifications. Trusts nothing without a unit test.
    ``\textit{Show me the latency numbers.}'' \\
  \bottomrule
\end{tabular}
\end{tcolorbox}

\section{Prompt for Synthetic Router Goals}\label{app:router_prompt}
\begin{center}
\begin{minipage}{0.95\linewidth}
\begin{tcolorbox}[title={Prompt for synthetic router goals},breakable,colback=white,colframe=black!60,boxrule=0.6pt,arc=2pt,left=6pt,right=6pt,top=6pt,bottom=6pt]
\begin{verbatim}
Target skill:
- name: {skill_name}
- description: {description}
- keywords: {keywords_block}

Task:
Generate synthetic router goals (queries) for this target skill.
The router state is ONLY a text goal (routing_goal).
Write realistic user-style goals.

Need:
- {need_pos} positive queries: target skill SHOULD be selected.
- {need_neg} hard negative queries: relevant to the same domain
  BUT target skill is not useful / not the best tool.

Hard negative requirements:
- Must look plausible and close to the target domain.
- Must share terminology/theme with the skill.
- Must be "relevant but useless" for THIS target skill.
- Avoid obvious cues like "do not use <skill>".

Style requirements:
- Do not mention the skill name directly.
- Keep each query concrete, actionable, and non-trivial.
- Mix concise and mildly noisy phrasing.
- English only (to match downstream tokenizer).

Already accepted positive queries (avoid duplicates):
{existing_pos_block}

Already accepted negative queries (avoid duplicates):
{existing_neg_block}

Return ONLY JSON in this schema:
{
  "positive_queries": [
    {"query": "...", "why_fit": "..."}
  ],
  "negative_queries": [
    {"query": "...", "why_relevant": "...", "why_useless": "..."}
  ]
}
\end{verbatim}
\end{tcolorbox}
\end{minipage}
\end{center}

\end{document}

%% file: macro.tex
\DeclareMathOperator*{\E}{\mathbb{E}}

\newcommand{\model}{\hat{p}}

\definecolor{blanchedalmond}{rgb}{1.0, 0.92, 0.8}
\definecolor{carmine}{rgb}{0.59, 0.0, 0.09}
\definecolor{lightblue}{rgb}{0.22,0.45,0.70}%

\newtheorem{theorem}{Theorem}[section]

\newtheorem{definition}[theorem]{Definition}

\renewcommand{\mathbf}{\boldsymbol}

\makeatletter
\def\Ddots{\mathinner{\mkern1mu\raise\p@
\vbox{\kern7\p@\hbox{.}}\mkern2mu
\raise4\p@\hbox{.}\mkern2mu\raise7\p@\hbox{.}\mkern1mu}}
\makeatother

\definecolor{amaranth}{rgb}{0.9, 0.17, 0.31}
\definecolor{antiquebrass}{rgb}{0.8, 0.58, 0.46}
\definecolor{antiquefuchsia}{rgb}{0.57, 0.36, 0.51}
\definecolor{chromeyellow}{rgb}{0.31, 0.47, 0.26}

